\documentclass[a4paper]{article}
\usepackage[text={7in,10in},centering]{geometry}
\usepackage{graphicx,color}
\usepackage{amsmath,amsthm,amssymb}
\usepackage{multirow,array}
\usepackage[table]{xcolor}
\usepackage{tabularx}  
\usepackage{booktabs}  
\usepackage{pifont}
\usepackage{tikz}
\usepackage{caption,subcaption}
\usepackage{url}
\usepackage{algorithm}
\usepackage[noend]{algpseudocode}
\usepackage{xurl}
\usepackage{authblk}
\usepackage{hyperref}
\usepackage[noabbrev]{cleveref}

\newtheorem{theorem}{Theorem}
\newtheorem{definition}{Definition}

\newcommand{\ones}{\mathbf 1}
\newcommand{\eg}{{\it e.g.}}
\newcommand{\ie}{{\it i.e.}}
\newcommand{\reals}{{\mbox{\bf R}}}

\newcommand{\minimize}{\mathop{\sf minimize{}}}
\newcommand{\argmin}{\mathop{\sf argmin}}

\newcommand{\wm}{\mbox{W/m}^2} 
\newcommand{\lpoint}{\mathcal{L}_{\text{Point}}}
\newcommand{\lpointk}{\mathcal{L}_{\text{Point}, k\text{-step}}}
\newcommand{\lpi}{\mathcal{L}_{\text{PI}}}
\newcommand{\lpik}{\mathcal{L}_{\text{PI}, k\text{-step}}}

\graphicspath{{figures/}}

\arrayrulecolor{teal}

\title{Barrier-enforced multi-objective optimization for direct point and sharp interval forecasting \footnote{This work has been submitted to the IEEE for possible publication. Copyright may be transferred without notice, after which this version may no longer be accessible.}} 

\author[1]{Worachit Amnuaypongsa}
\author[1]{Yotsapat Suparanonrat}
\author[2]{Pana Wanitchollakit}
\author[1]{Jitkomut Songsiri\thanks{Corresponding author.}}

\affil[1]{Department of Electrical Engineering, Faculty of Engineering, Chulalongkorn University, Bangkok, Thailand}
\affil[2]{Department of Computer Engineering, Faculty of Engineering, Chulalongkorn University, Bangkok, Thailand}

\affil[ ]{\textit{e-mail: worachitam@gmail.com, YotsapatSK140@gmail.com, pnaw.wan@gmail.com, jitkomut.s@chula.ac.th}}

\begin{document}
\maketitle

\begin{abstract}
This paper proposes a multi-step probabilistic forecasting framework using a single neural-network based model to generate simultaneous point and interval forecasts. Our approach ensures non-crossing prediction intervals (PIs) through a model structure design that strictly satisfy a target coverage probability (PICP) while maximizing sharpness. Unlike existing methods that rely on manual weight tuning for scalarized loss functions, we treat point and PI forecasting as a multi-objective optimization problem, utilizing multi-gradient descent to adaptively select optimal weights. Key innovations include a new PI  loss function based on an extended log-barrier with an adaptive hyperparameter to guarantee the coverage, a hybrid architecture featuring a shared temporal model with horizon-specific submodels, and a training strategy. The proposed loss is scale-independent and universally applicable; combined with our training algorithm, the framework eliminates trial-and-error hyperparameter tuning for balancing multiple objectives. Validated by an intra-day solar irradiance forecasting application, results demonstrate that our proposed loss consistently outperforms those in current literature by achieving target coverage with the narrowest PI widths. Furthermore, when compared against LSTM  encoder-decoder and Transformer architectures--including those augmented with Chronos foundation models--our method remains highly competitive and can be seamlessly adapted to any deep learning structure.
\end{abstract}

\section{Introduction}
\label{sec:introduction}


In time series forecasting applications for a variable $y$, predictions are generally categorized into point forecasts and interval forecasts. A point forecast, denoted as $\hat{y}$, provides a single estimate for a future time step, typically representing the most likely value or the mean of the predictive distribution. In contrast, an interval forecast--commonly as a form of prediction interval (PI) quantifies the uncertainty of the estimate. It is defined by a range $[\hat{l}, \hat{u}]$ that is expected to contain the true future value of $y$ with a pre-specified confidence level. In renewable energy (RE) applications, the random nature of RE poses significant challenges to grid stability and energy management system efficiency. Point forecasts are used across multiple timescales in unit commitment and economic dispatch to maintain the balance between supply and demand while ensuring grid stability and economic efficiency. Shifting to risk-based decision making, PIs provide the necessary bounds for determining operating reserve, and robust optimizations in grid management \cite{osti_1060669, Zhao2022, Amnuaypongsa2025}. The cost of point and probabilistic forecast errors is primarily driven by the magnitude and sign of the deviation alongside the price spread between markets, as well as the PI quality. These inaccuracies lead to significant grid operational expenses, inefficient dispatch, and the economic burden of maintaining reserves to cover uncertainty \cite{gandhi2024forecasterror}.


Given the importance of PI estimation, an extensive literature on quantile regression (QR) \cite{Koenker2017quantile} and QR neural network (QRNN) variants using pinball loss has emerged; however, producing high-quality intervals remains challenging due to the inherent trade-off between reliability (how well the PI covers the actual samples) and sharpness (how \emph{tight} the interval is). These two metrics are typically measured by the Prediction interval coverage probability (PICP) and the PI width. These objectives are naturally conflicting: a high PICP--indicating superior coverage--often results in a wider interval. In practice, excessive width leads to over-conservatism in uncertainty-aware system planning; in power systems, this incurs higher operational costs as more resources must be reserved for worst-case scenarios. To address this, several studies propose specialized loss functions to balance these goals by penalizing PI width. Commonly, PI loss functions merge PICP and width into a scalar value, such as the multiplicative CWC \cite{Khosravi2011} or additive DIC \cite{Zhang2015}. Recent work has shifted toward gradient-based formulations like the likelihood-based QD loss \cite{Pearce2018}, which uses smooth PICP approximations. This framework has been refined in recent variants \cite{Saeed2024}. Reducing PI over-conservatism is a primary focus of \cite{probforecastPMAP24}, which introduced a large-width penalty using the sum-of-$K$ largest function. This was evolved into the Sum-$k$ loss \cite{Amnuaypongsa2025} used with NN-based models to achieve even tighter intervals. In contrast to the QRNN approach, these methods explicitly incorporate PI width as an optimization objective rather than a secondary outcome. Nevertheless, they require hyperparameters to balance PICP and PI width. Since achieving the target PICP is the priority, performance depends on carefully tuning these parameters. 


Risk-aware decision-making requires both point forecasts and uncertainty quantification through PIs. Consequently, research has shifted toward integrated NN-based models capable of generating both outputs simultaneously. Literature in this direction proposed a loss function that merges point and PI objectives into a single value.  Based on quantile loss, \cite{Kivaranovic20a} provided a framework for NNs to produce a point prediction and valid PI by leveraging a quantile-based approach and split conformal prediction. While the paper addresses valid coverage, it does not fully optimize for maintaining the narrowest possible intervals. Several approaches introduced complex loss functions that require careful hyperparameter tuning to balance competing objectives. For example, extended from QD loss, the QD+ loss \cite{Pearce2018} added MSE as a point loss and a penalty for coverage violations. IPIV \cite{Simhayev2022} constructed point forecasts via a convex combination of PI bounds, using a weighted sum of PI and regression objectives, which balances PICP and width via a weight parameter. A midpoint approach \cite{Lai2022} defined the point forecast as the PI midpoint and replaced the Gaussian log-likelihood variance term with a width penalty, though it lacks explicit PICP guarantees and requires extra tuning. Further complexity is introduced in the EMVE loss  \cite{Saeed2025a} which combined modified pinball loss, ordering quantile penalties, and coverage/width terms to address extreme tail estimation, significantly increasing complexity. Finally, the EMVE loss \cite{Saeed2025b} integrates width, Gaussian likelihood, and quantile-violation penalties; however, because these terms use disparate units and rely on a Gaussian assumption, the framework may not generalize well to non-Gaussian data. Other methodologies focused on specific neural network architectures or training strategies. For instance, PI3NN \cite{liu2022pi3nn} employed three separate networks for $\hat{l}, \hat{y}$, and $\hat{u}$--each trained independently with MSE--which increases model complexity and computational overhead. Alternatively, DualAQD \cite{Morales2023} utilized a two-stage process where a point forecasting model is pre-trained before estimating PIs using an adaptively weighted sum of width and coverage penalties. However, because these loss terms are not properly scaled, controlling their relative importance during training remains a challenge.


In a general context, forecasting quality depends on model architecture, the loss function, training mechanisms, and input selection. For multi-step point and probabilistic forecasting, the primary objective is producing a valid PI that covers the point prediction and achieves the target PICP, as insufficient coverage--such as in RE reserve preparation--can lead to a higher deficit penalty like the value of lost load than other penalties \cite{Zhao2021}. Once the PICP target is reliably met, the secondary goals are to minimize PI width to reduce over-conservatism and maintain acceptable deterministic forecasts. Balancing these goals across multi-step horizons is particularly challenging, as interval width naturally expands with increasing lead times. While existing literature addresses these goals to some extent, the main challenge remains the simultaneous optimization of multiple objectives. Using a scalarized loss requires a proper balance of weights, which becomes difficult to tune when objective terms overlap in function or are not properly scaled. 

This paper addresses these critical points by proposing the following contributions:

\begin{enumerate}
\item \textbf{Structural enforcement of valid PIs:} Unlike existing literature that relies on complex penalty terms to prevent bound crossing or to force $\hat{y}$ within the PI, we structurally design the network output layer to satisfy these constraints. This architectural approach ensures valid PIs by construction and reduces the need for complex penalties.
\item \textbf{Barrier-enforced composite loss:} We propose a novel loss function combining regression loss for point estimation, PI width penalties, and a new PICP loss. The latter utilizes an extended log-barrier function to strictly penalize coverage violations. All terms are designed to be scale-independent for universal applicability.
\item \textbf{Multi-objective training algorithm:} By treating point and PI estimation as dual objectives, we apply a training algorithm that identifies a common gradient descent direction to \emph{mutual improvement} in both objectives. This approach eliminates the weight tuning typically required by scalarized loss functions in existing literature.
\item \textbf{Seamless architecture integration:} The proposed smooth loss and training mechanisms are easily integrated into advanced deep learning frameworks, demonstrated by augmenting foundation models into tailor-made probabilistic forecasting architectures.
\end{enumerate}

These improvements are integrated into our proposed framework, \textbf{SolarPointPI}, consisting of a horizon-specific NN architecture, a novel loss function, and a targeted training strategy. While the framework is universal, we demonstrate its efficacy through solar irradiance forecasting, with implementation details tailored specifically for this application. 


Existing literature on point and interval forecasting of solar irradiance often lacks a direct mechanism for constructing PIs as primary model outputs or incorporating PI widths as a design objective. Most frameworks generate PIs through a secondary stage: for instance, \cite{Fatemi2018probsolar} parameterized Beta distributions to estimate CDFs, while others rely on post-processing techniques like KDE on XGBoost forecasts \cite{Li2022probsolar}, Gaussian processes \cite{Huang2020probsolar}, or data clustering with $k$-means \cite{Scolari2016probsolar}. Even modern hybrid architectures that merge statistical methods with NNs, such as those in \cite{Gneiting2023probsolar} and \cite{Singla2023probsolar}, typically focus on quantile production rather than width control. This two-stage approach--seen in \cite{Wang2017probsolar}, where deep CNN point forecasts are subsequently converted into quantiles--results in sharpness being an outcome of the statistical distribution rather than a primary design constraint. While advancements in solar energy forecasting could alternatively rely on complex input selection, or high-capacity deep learning models \cite{paletta2023advances} requiring vast datasets, our framework focuses on practical improvements of PI quality that bypass the prohibitive costs of extensive data requirements.

\section{Methodology}
\label{sec:methodology}

This section presents the methodology for generating point and PI forecasts in a multi-step-ahead forecasting framework. It consists of three parts: a model architecture, a loss function, and a training algorithm. The model architecture explains how the point forecast and its corresponding upper and lower bounds are structured to ensure that the point estimate remains within the PI. A loss function is proposed to firmly achieve PIs with a desired coverage probability and with a special modification for solar irradiance data that have a diurnal cycle. The training algorithm details how the multi-objective optimization problem is addressed using the multi-gradient descent algorithm (MGDA).

\subsection{Model architecture}
\label{subsec:model-architecture}

\begin{figure}
\centering
\includegraphics[width=0.5\linewidth]{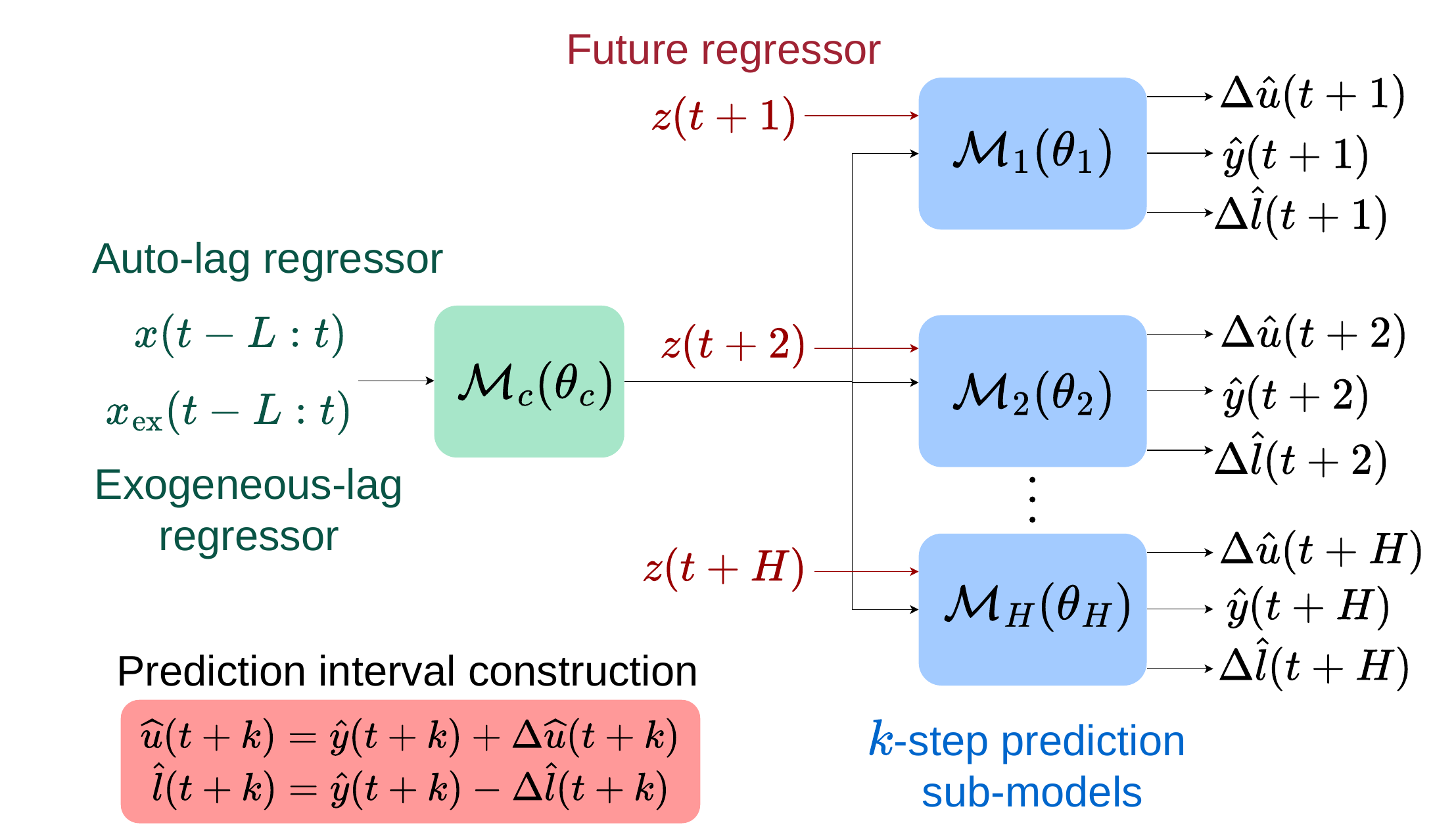}
\caption{A proposed point and PI forecasting model architecture.}
\label{fig:pipoint_model}
\end{figure}

The proposed model is designed to jointly produce the upper bound $\hat{u}$, lower bound $\hat{l}$, and point forecast $\hat{y}$ for a forecasting horizon $H$, ensuring that the point forecast always lies within the PI. Let $x$ denote auto-lag regressors, $x_{\text{ex}}$ exogenous lag regressors, $z$ future regressors, and $y$ the target vector (for multi-step outputs). \Cref{fig:pipoint_model} illustrates that the model consists of a \textbf{common model} $\mathcal{M}_{c}(\theta_{c})$ and \textbf{submodels} $\mathcal{M}_{k}(\theta_{k})$ with their model parameters, $\theta_{c}$ and $\theta_k$ for $k = 1, \ldots, H$. The common model, which can be any neural network model, is shared across all forecasting steps and processes historical information from the lagged inputs, $x$ and $x_{\text{ex}}$. Each submodel, on the other hand, specializes in a specific forecast horizon, taking as input the hidden representation from the common model together with the future regressor $z$ corresponding to its prediction step. For a $k$-step forecast, the $k^{\text{th}}$ submodel releases three quantities: $\hat{y}$, $\Delta \hat{u}$, and $\Delta \hat{l}$, and we can construct the PI according to $\hat{u} = \hat{y} + \Delta \hat{u}$ and $\hat{l} = \hat{y}- \Delta \hat{l}$. 

To ensure that the point forecast $\hat{y}$ remains within the PI, we enforce $\Delta \hat{u}$ and $\Delta \hat{l}$ to remain non-negative using the Softplus activation function, which is strictly positive and differentiable, to avoid the zero-gradient issue and ensure proper gradient flow for the PICP objective. A linear activation is retained for the point forecast $\hat{y}$.

The proposed model in \Cref{fig:pipoint_model} is flexible and can be served for any forecasting application, allowing users to substitute any model architecture for $\mathcal{M}_{c}(\theta_{c})$ and $\mathcal{M}_{k}(\theta_{k})$. In our experiments, we find that using an LSTM in $\mathcal{M}_{c}(\theta_{c})$ to capture temporal dynamics and a feed-forward neural network in $\mathcal{M}_{k}(\theta_{k})$ is sufficient to produce desirable results in our setting. We introduce the \textbf{SolarPointPI} framework, which integrates the proposed model with a new PI and point loss function, and a multi-objective training algorithm - all detailed in subsequent sections, to jointly produce PI and point forecasts.

\subsection{Problem formulation}
\label{subsec:problem-formulation}

In this study, we define two learning objectives: producing high-quality point forecasts and constructing reliable PIs, optimized via a point forecast loss (regression loss) and PI loss, respectively. We denote the model parameters as $\theta = (\theta_c, \theta_1,\ldots,\theta_H)$. 
Conceptually, the point forecast $\hat{y}$ and PI bounds ($\hat{l}, \hat{u}$) are functions of $\theta$, indexed by time $t$ (as sample index), and $k=1,2,\ldots,H$ for a $k$-step prediction. Denoted as $\hat{y}(t+k|t; \theta)$, these outputs are calculated at time $t$ for the future time $t+k$. Consequently, loss functions are typically aggregated across both samples and prediction steps.

\subsubsection*{Point estimation loss} 
The point estimation loss can be employed using any standard regression loss $\ell$ and is normalized by data range $R_{Q}$, the number of samples $N$, and the number of prediction steps. 
\begin{equation}
\lpoint(\theta) = \frac{1}{H} \sum_{k=1}^{H} \lpointk(\theta) = \frac{1}{HNR_{Q}}\sum_{k=1}^{H} \sum_{t =1}^N \ell ( y(t+k), \hat{y}(t+k|t;\theta)).
\label{eq:pointloss_allsteps}
\end{equation}
The $R_Q$ normalization factor is is defined as the interquantile range $R_{Q} = q_{y}(0.95) - q_{y}(0.05)$, rendering the loss function unitless and independent of variable magnitude. Instead of using min-max range, $R_Q$ is less sensitive to outliers. This work applies the MAE loss, $\ell(y,\hat{y}) = |y-\hat{y}|$ as it offers greater robustness to outliers compared to the standard MSE loss.

\subsubsection*{PI estimation loss}
The primary objective of PI estimation is to guarantee the target PICP across multi-step horizons, followed by minimizing interval width. In RE applications, ensuring this coverage is particularly challenging when managing reliability to avoid reserve deficit penalties \cite{Zhao2021}. To address this, let the PICP deviation from a target probability $p$ be $z = p - \text{PICP}$, which is desired to be non-positive. Constrained optimization often enforces the requirement $z \leq 0$ using the log-barrier function $\psi_r(z) = -\frac{1}{r}\log(-z)$, which requires $z$ to be strictly negative. However, as a PI loss, the log-barrier is undefined for $z \ge 0$, rendering the loss ill-defined if the model is initialized in an infeasible region. To resolve this issue while preserving a strong penalty on the PICP deviation, we propose to use the \textbf{extended log-barrier} penalty that strongly penalizes $z$.

\begin{equation} 
\psi_{r}(z) =
\begin{cases}
-\frac{1}{r}\log(-z), & \text{if } z \leq -\frac{1}{r^2},\\
rz - \frac{1}{r}\log\!\left (\frac{1}{r^{2}}\right ) + \frac{1}{r},
& \text{otherwise}.
\end{cases}
\label{eq:logbarrier}
\end{equation}

\begin{figure}
\centering
\begin{subfigure}[b]{0.66\textwidth}
\centering
\includegraphics[width=\linewidth]{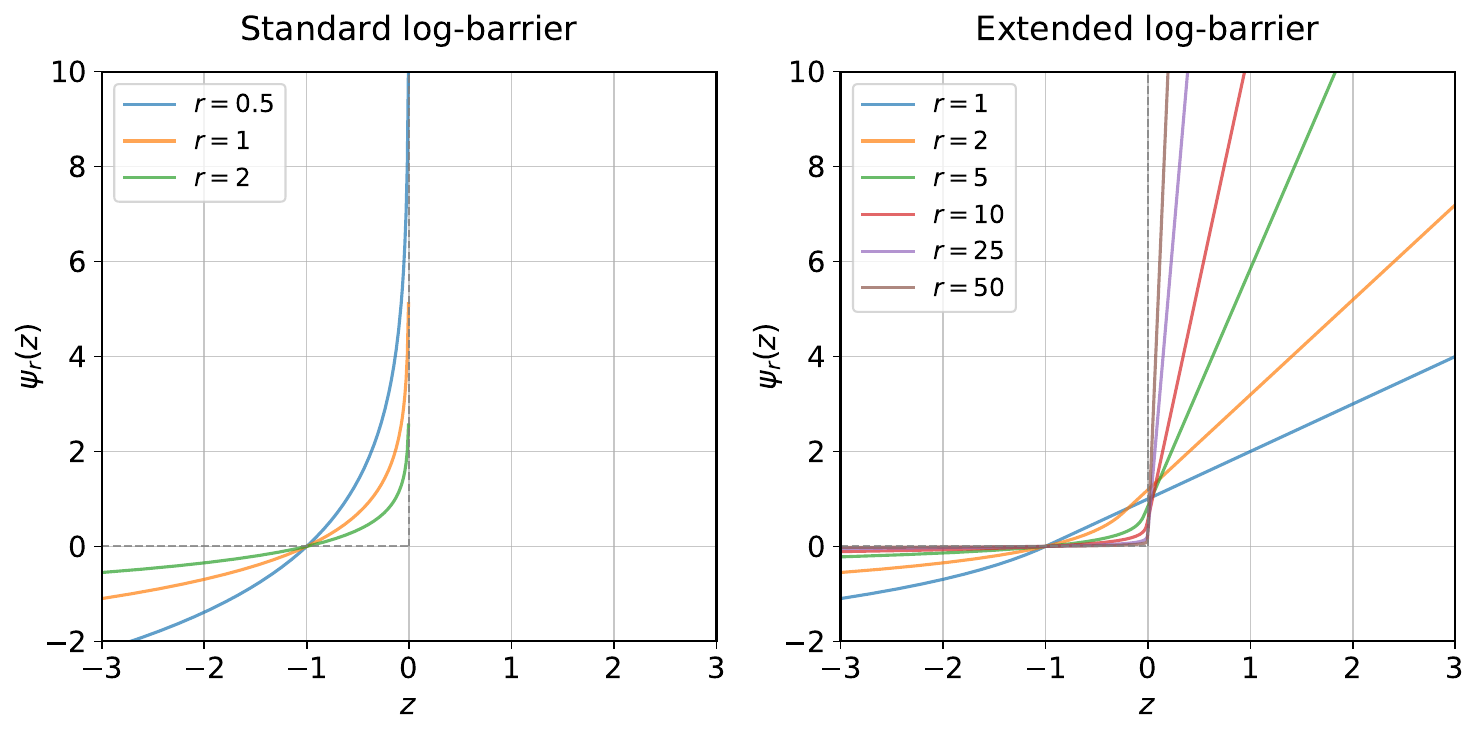}
\caption{Illustration of the extended log-barrier function $\psi_{r}(z)$.}
\label{fig:logbarrierfunction}
\end{subfigure}
\hfill
\begin{subfigure}[b]{0.33\textwidth}
\centering
\includegraphics[width=\linewidth]{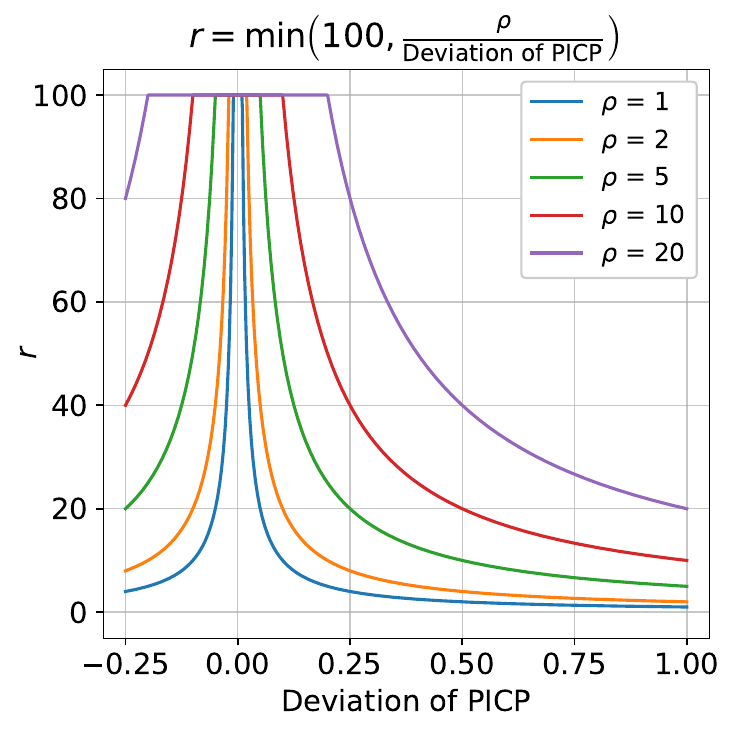}
\caption{Illustration of adaptive $r$ formulation.}
\label{fig:adaptive_r}
\end{subfigure}
\caption{Penalty function components showing the extended log-barrier and the adaptive parameter $r$.}
\label{fig:combined_barrier_plots}
\end{figure}

The extended log-barrier function with parameter $r > 0$ was used in \cite{Kervadec2022} in the context of learning deep neural networks with constraints. The logarithm term in the standard log-barrier is replaced with a first-order Taylor expansion beyond a threshold $-1/r^2$, yielding a function that is strictly convex, twice differentiable, and defined for all $z \in \reals$, thereby guaranteeing continuous gradients throughout training even when the coverage constraint is violated. The hyperparameter $r$ acts as a barrier softening factor; the larger, the higher penalty on the infeasible region as shown in \Cref{fig:logbarrierfunction}. To dynamically tighten the penalty boundary and increase the penalty on the model when the target coverage $p_k$ is far from the desired PICP, we calculate $r$ adaptively and inversely proportional to the PICP deviation: 
\begin{equation} 
r = \min\left ( 100, \frac{\rho}{\left | p - \text{PICP} \right|}\right ),
\label{eq:adaptive_r}
\end{equation}
where a multiplier factor $\rho$ is empirically set to $10$ to ensure a stable and reliable optimization process (\Cref{fig:adaptive_r}). This adaptive mechanism allows $r$ to remain small when the model is far from the target, effectively softening the extended log-barrier function $\psi_r(z)$ (\Cref{fig:adaptive_r}a). This prevents aggressive gradients from forcing parameters too abruptly during early training. As the PICP approaches the target, $r$ increases toward its cap of $100$, hardening the penalty boundary to strictly enforce the coverage constraint. Consequently, the coverage penalty $\psi_r (p_k - \text{PICP}_k(\theta) )$ is applied across each $k$-step prediction to ensure reliable interval estimation.

For PI width penalty, denote an interval width as $w(\theta) = \hat{u}(\theta) - \hat{l}(\theta)$ (indexed by samples and $k$-step predictions) and is a function of $\theta$. We apply the concept of the Sum-$k$ loss function \cite{Amnuaypongsa2025} which penalizes the sum of the $k$-largest interval widths. This approach effectively reduces extremely large intervals, mitigating over-conservatism when using PI bounds in robust optimization. Defining this width penalty begins with specifying a hyperparameter $K$ (positive integer) as the number of \emph{large} width samples, such as $K = \lfloor (0.3N) \rfloor$ meaning that the top $30\%$ of width samples will be penalized heavily. Then, $w_{[i]}$ denotes the $i$-th largest width according to $w_{[1]} \geq w_{[2]} \geq \cdots \geq w_{[N]}$. The Sum-$k$ loss penalizes the large-width samples and the remaining smaller-width samples in different degrees controlled by a hyperparameter $\lambda \geq 0$. The width penalty in the Sum-$k$ loss is described by
\begin{equation} 
\mathcal{W}(\theta) = \frac{1}{R_{Q}} \left [ \frac{1}{K} \sum_{i=1}^{K} w_{[i]}(\theta)
        + \frac{\lambda}{N-K} \sum_{i=K+1}^{N} w_{[i]}(\theta)
    \right ].
\label{eq:individual_width}
\end{equation}
Setting $\lambda < 1$, such as 0.8 in our experiment, means that we penalize more heavily on the average large width as compared to the average of smaller widths. The normalization term $R_{Q}$ scales the loss relative to the target variable range, making $\mathcal{W}$ unitless and ensuring comparability across different datasets. While the sum-of-$k$-largest function is non-differentiable, frameworks like PyTorch handle it effectively using a subgradient approach.

\paragraph{Proposed PI loss.} Combining the PI coverage and PI width penalty described above, the overall PI loss across all $H$ forecasting horizons is defined as the average of the $k$-step prediction loss
\begin{equation} 
\lpi(\theta) = \frac{1}{H}  \sum_{k=1}^H \lpik (\theta) = \frac{1}{H}  \sum_{k=1}^H \psi_r (p_k - \text{PICP}_{k}(\theta) ) + \mathcal{W}_k(\theta).
\label{eq:piloss_allsteps}
\end{equation}
Each term in $\lpi$ is unitless and scale-independent. It requires only one user-defined hyperparameter: $\lambda$ in \eqref{eq:individual_width}.

\paragraph{Smooth approximation of calculating PICP.} Conventionally, the standard counting function, $\ones \{ y \in [\hat{l},\hat{u}] \}$ is used to compute PICP. To preserve differentiability for backpropagation, it is replaced with a smooth \texttt{tanh} approximation:
\begin{equation}
\text{PICP} = \frac{1}{N} \sum_{i=1}^N \ones_{\tanh} \{ y_i \in [\hat{l}_i,\hat{u}_i] \}, \label{eq:picp} 
\end{equation}
where $\ones_{\tanh} \{  y \in [\hat{l},\hat{u}] \}
= (1/2) \max \{ 0, \tanh \big(s(y - \hat{l})\big)
        + \tanh \big(s(\hat{u} - y)\big) \} $,
and $s > 0$ is a smoothing factor.
\paragraph{PI coverage loss for solar energy.} 
In solar irradiance forecasting, irradiance is zero at night. Enforcing uniform coverage across all hours complicates the trade-off between interval width and coverage, as nighttime zeros are trivially covered even by wide, unoptimized PIs. To address this, we apply the extended log-barrier function separately to daytime and nighttime samples, assigning distinct target coverage levels ($p$): $0.90$ for day and $0.15$ for night. Samples are split into day and night sets by directly thresholding irradiance values (\eg, $0.001$), bypassing the need for date-time metadata in the training process. We adjust the PICP by applying a smooth mask function, $\max(0, \tanh(\cdot))$, that returns a mask value in $[0,1]$ to isolate daytime and nighttime samples:
\begin{align}
\text{PICP}_{\text{day}}
&= \frac{1}{N_{\text{day}}} \sum_{i=1}^{N} \ones_{\tanh} \{ y_i \in [\hat{l},\hat{u}] \} \cdot \max \bigl \{0,\; \tanh  ( s (y_{i}
           - \text{threshold}  )  )\bigr \} , \label{eq:picp_day}\\
\text{PICP}_{\text{night}}
& = \frac{1}{N_{\text{night}}} \sum_{i=1}^{N} \ones_{\tanh} \{ y_i \in [\hat{l},\hat{u}] \} \cdot \max \bigl \{ 0,\; \tanh ( s (\text{threshold}- y_{i}  ) ) \bigr \},
\label{eq:picp_night}
\end{align}%
where $N_{\text{day}} = \sum_{i=1}^N \max \{0,\; \tanh  ( s (y_{i} - \text{threshold}  )  ) \}$ and $N_{\text{night}} = \sum_{i=1}^N \max  \{ 0,\; \tanh ( s (\text{threshold}- y_{i}  ) ) \}$. 
Accordingly, the extended log-barrier penalty for each $k$-step in \eqref{eq:piloss_allsteps} is split into separate day and night terms: $\psi_r\!(  0.90 - \text{PICP}_{\text{day}} ) 
    + \psi_r\! (  0.15 - \text{PICP}_{\text{night}} )$. 

We conclude this section by defining the goal of model learning as the simultaneous minimization of two distinct objectives: the point and PI estimation losses as defined in \eqref{eq:pointloss_allsteps} and \eqref{eq:piloss_allsteps}, respectively. Traditional approaches merge multiple goals into a single value as minimizing $\gamma_1 \lpoint + \gamma_2 \lpi$, while selecting a value of weights $\gamma=(\gamma_1,\gamma_2)$ yields a Pareto optimal solution. This leaves the user to find a meaningful balance between two objectives that often results in a trial and error process. Unlike this direction, we treat this task strictly as a multi-objective optimization (MOO) problem. Formally, we seek to optimize the vector-valued loss function:
\begin{equation} 
\minimize_{\theta} \ \mathbf{L}(\theta) = \left( \lpoint(\theta), \lpi(\theta) \right).
\label{eq:twoobj_optim}
\end{equation}
By adopting the MOO framework instead of scalarization, we avoid the biases inherent in fixed weighting. This approach enables the model to more effectively explore the trade-offs between point estimation accuracy and PI quality. The following section introduces a framework designed to achieve this balance.

\subsection{Training algorithm}
\label{subsec:mgda}
We treat the proposed problem as an MOO task and first outline its fundamental optimality. We then utilize the multiple gradient descent algorithm (MGDA) \cite{desideri2012MGDA, Sener2018} to identify a common descent direction that simultaneously improves all objectives. This is achieved by dynamically calculating weights at each iteration based on the local geometry of the loss gradients. Finally, we summarize the training algorithm resulting from applying these concepts to our specific problem~\eqref{eq:twoobj_optim}.

\subsubsection*{Multi-objective optimization}
We first introduce the fundamental concepts of MOO. Consider $m$ tasks, each associated with an objective function $\mathcal{L}_i: \reals^n \rightarrow \reals$ for $i \in [m]$ where $[m]$ denotes the index set $\{1, 2, \ldots, m\}$.  Unlike single-objective optimization, solutions in MOO cannot be ranked using a single scalar criterion; instead, they are evaluated through the concept of dominance. The concept of optimality can be defined as follows.

\begin{definition}
\cite{miettinen1999} A solution $\theta^{(a)}\in \reals^n$ is said to be \textbf{dominated} by another solution $\theta^{(b)} \in \reals^n$ iff $\mathcal{L}_i(\theta^{(a)} \leq \mathcal{L}_i(\theta^{(b)})$ for all $i \in [m] $ and there exists at least one $i \in [m]$ such that $\mathcal{L}_i(\theta^{(a)} < \mathcal{L}_i(\theta^{(b)})$. A solution $\theta^{\star}$ is said to be \textbf{Pareto optimal} if it is not dominated by another solution. A \textbf{Pareto set} is the set of all Pareto optimal solutions. A \textbf{Pareto front} is the set of all objective function values of the Pareto optimal solutions. 
\end{definition}
While gradient-based methods in deep learning typically target stationary points for single-objective problems, MOO requires a Pareto stationary definition. Denote $\Delta_{m}$ the simplex set: $\{ a \in \reals^m | \sum_{i=1}^M a_i =1, a_i \geq 0 , i \in [m] \;\}$.
\begin{definition}
\cite{desideri2012MGDA} The objective functions $\mathcal{L}_i$ for all $i \in [m]$ are said to be \textbf{Pareto-stationary} at $\theta^\star$ iff there exists a convex combination of the gradients, that is equal to zero: $\sum_{k=1}^m \gamma_k \nabla_\theta \mathcal{L}_k(\theta^\star) = 0$ where $\gamma \in \Delta_m$.
\end{definition}
Pareto stationary is a necessary condition for Pareto optimality. If all $\mathcal{L}_i$ are convex with $\gamma_i > 0,\;i \in [m]$, it is the Karush-Kuhn-Tucker (KKT) sufficient and necessary condition for Pareto optimality \cite{desideri2012MGDA}. 

While many studies in MOO or multi-task learning aim to identify the entire Pareto front, our objective in \eqref{eq:twoobj_optim} is more targeted. We seek a Pareto optimal solution specifically within the high-reliability region where the PICP is high. Consequently, we focus on a neighborhood of the target $\lpi$ value rather than the complete front. Research directions in MOO in finding a single Pareto optimal solution generally fall into two categories: scalarization via fixed weights or dynamic weight learning \cite{chen2025reviewMOO}. Among gradient-balancing methods, we select MGDA because it addresses both magnitude and directional conflict between competing objectives. Furthermore, its inherent scale-invariance eliminates the trial-and-error process of hyperparameter tuning.

\subsubsection*{Multiple gradient descent algorithm}
MGDA extends the steepest gradient descent to MOO and is proved to converge to a Pareto stationary point. The main result is given as follows. 
\begin{theorem} \cite{desideri2012MGDA} Consider the convex hull of the objective gradients given by $\mathcal{C} = \{ u \in \reals^n \;|\; u = \sum_{k=1}^m \gamma_k \nabla \mathcal{L}_k(\theta), \; \gamma \in \Delta_m\; \text{(the simplex)} \;\}$. Let $g$ be the \textbf{minimum-norm} element of $\mathcal{C}$ at $\theta_0$. Then we have two possible outcomes: i) $g = 0$ and hence, $\mathcal{L}_i(\theta)$ for $i \in [m]$ are Pareto-stationary at $\theta_0$, and ii) when $g \neq 0$ then $-g$ is a descent direction \textbf{common} to all the objectives, \ie, $\langle \nabla \mathcal{L}_i, -g  \rangle \leq 0,\;i \in [m]$. Additionally, if $g$ is interior to $\mathcal{C}$, it can be shown that $\langle \nabla \mathcal{L}_i, g \rangle = \Vert g \Vert^2, \forall i \in [m]$.
\label{thm:mgda}
\end{theorem}
From \Cref{thm:mgda}, if we choose our search direction as $-g$, specifically from the minimum-norm element in $\mathcal{C}$, this does not only provide any common descent direction, the improvement for any objective $i$: $\mathcal{L}_i(\theta)-\mathcal{L}_i(\theta-\eta g) \approx -\eta \nabla \mathcal{L}_i^T g$ is at least $-\eta \Vert g \Vert^2$ where $\eta$ is the step size. It is an optimal descent direction in the sense of fairness. The algorithm ensures that the task with the least favorable gradient (the one most in conflict with others) still receives possible improvement allowed by the geometry.


MGDA requires solving the minimum-norm problem to obtain the optimal weight $\gamma$:
\begin{equation}
\gamma^\star = \argmin_\gamma \;\; \left \Vert  \sum_{k=1}^m \gamma_k \nabla \mathcal{L}_k (\theta) \right \Vert_2^2 \quad \text{subject to} \quad \gamma \in \Delta_m \;\text{(the simplex)}.
\label{eq:minnorm}
\end{equation}

The problem is to find the shortest vector within the convex hull of objective gradients. If $g =\sum_{k=1}^m \gamma_k^\star \nabla \mathcal{L}_k (\theta) \neq 0$ then $-g$ is a descent direction that improves all objectives. For $m$ objectives where $m \geq 2$, the problem \eqref{eq:minnorm} is a quadratic program where \cite{Sener2018} use the Frank-Wolfe algorithm to solve it. For two objectives, the problem~\eqref{eq:minnorm} reduces to a one-dimensional problem:
\begin{equation}
\gamma_1^\star = \argmin_{a} \;\; \Vert a \nabla \mathcal{L}_1 + (1-a) \nabla \mathcal{L}_2 \Vert_2^2 \;\; \text{subject to} \;\; 0 \leq  a \leq 1, \quad \gamma_2^\star = 1 - \gamma_1^\star.
\label{eq:minnorm2obj}
\end{equation}
The objective of \eqref{eq:minnorm2obj} is $a^2 \Vert \nabla \mathcal{L}_1 - \nabla \mathcal{L}_2\Vert^2 + 2a (\nabla \mathcal{L}_1 - \nabla \mathcal{L}_2)^T \nabla \mathcal{L}_2  + \Vert \nabla \mathcal{L}_2 \Vert_2^2$. The zero-gradient condition with respect to $a$ is $a_0 = \frac{(\nabla \mathcal{L}_2 - \nabla \mathcal{L}_1)^T\nabla \mathcal{L}_2}{\Vert \nabla \mathcal{L}_1 - \nabla \mathcal{L}_2\Vert^2} $. Therefore, with constraint $0 \leq a \leq 1$, the optimal $a^\star$ is obtained by clipping $a_0$ onto $[0,1]$.
\begin{itemize}
\item Case 1: $a_0 \leq 0 \Leftrightarrow \langle \nabla \mathcal{L}_1, \nabla \mathcal{L}_2 \rangle \geq \Vert \nabla \mathcal{L}_2 \Vert^2 \Leftrightarrow \Vert \nabla \mathcal{L}_1 \Vert \cos \theta \geq \Vert \nabla \mathcal{L}_2 \Vert $. This gives $a^\star = 0, \gamma_1^\star = 0, \gamma_2^\star = 1$.
\item Case 2: $0 < a_0 < 1 \Leftrightarrow \langle \nabla \mathcal{L}_1, \nabla \mathcal{L}_2 \rangle \leq \Vert \nabla \mathcal{L}_1 \Vert^2$ and $  \langle \nabla \mathcal{L}_1, \nabla \mathcal{L}_2 \rangle \leq \Vert \mathcal{L}_2 \Vert^2 $. Hence, $a^\star = a_0, \gamma_1^\star = a_0, \gamma_2^\star = 1- \gamma^\star$.
\item Case 3: $a_0 \geq 1 \Leftrightarrow \langle \nabla \mathcal{L}_1, \nabla \mathcal{L}_2 \rangle \geq \Vert \nabla \mathcal{L}_1 \Vert^2 \Leftrightarrow \Vert \nabla \mathcal{L}_2 \Vert \cos \theta \geq \Vert \nabla \mathcal{L}_1 \Vert $. This gives $a^\star = 1, \gamma_1^\star = 1, \gamma_2^\star = 0$.
\end{itemize}
In conclusion, the analytical solution to \eqref{eq:minnorm2obj} can be compactly expressed as
\begin{figure}
 \centering
 \includegraphics[width=0.5\linewidth]{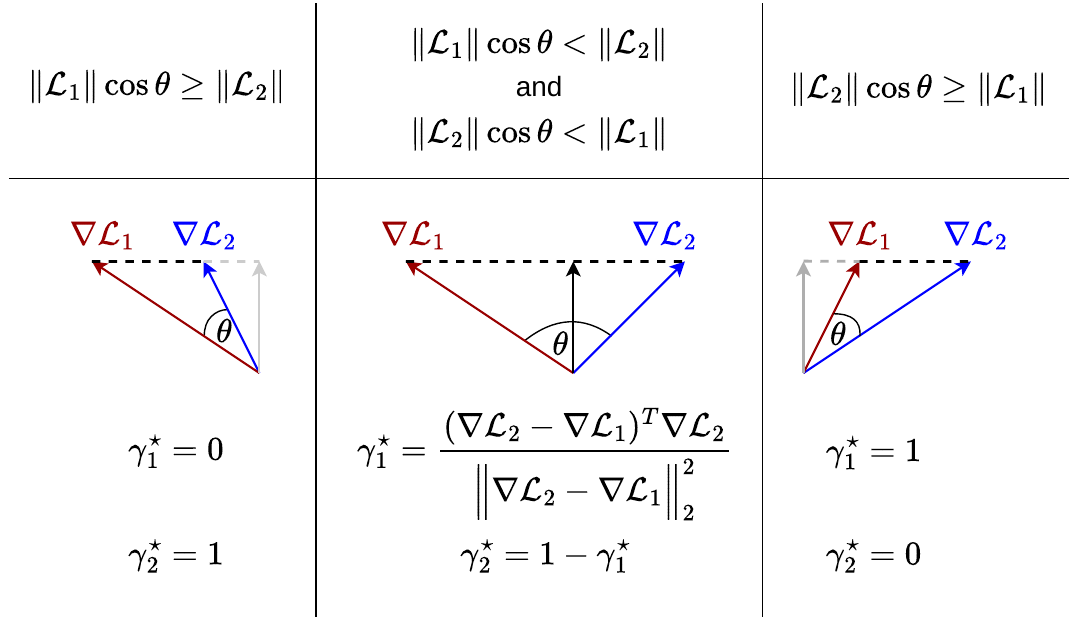}
  \caption{Geometric interpretation of the minimum-norm problem \cite{Sener2018}. The optimal weights $\gamma_1^*$ and $\gamma_2^*$ identify the shortest vector in the convex combination (black dashed line) formed by $\nabla \mathcal{L}_1$ and $\nabla \mathcal{L}_2$. The faint grey arrows represent the shortest-norm vector that is a linear combination of two gradients when there is no constraint on $\gamma_1+\gamma_2 =1$.}
 \label{fig:mgdasol_twoobj}
\end{figure}

\begin{equation} 
 \gamma_1^{\star} = \max \left (  \min  \left (
   \frac{(\nabla\mathcal{L}_2 - \nabla\mathcal{L}_1)^T \nabla \mathcal{L}_2}
        { \Vert \nabla\mathcal{L}_2 - \nabla\mathcal{L}_1 \Vert_2^2},
   \; 1 \right ), \; 0  \right ), \;\; \gamma_2^\star = 1-\gamma_1^\star.
\label{eq:closedform_gamma}
\end{equation}
This solution is illustrated geometrically in \Cref{fig:mgdasol_twoobj}. If the gradient of task $i$ is already contained in the direction of task $j$ (examined by $\Vert \nabla \mathcal{L}_j \Vert \cos \theta \geq \Vert \nabla \mathcal{L}_i \Vert$), then by following the descent direction of task $i$, we are guaranteed to decrease loss $j$ as well. The optimal weight is then assigned entirely to task $i$ which provides the most efficient common descent at that specific geometry. Conversely, when gradients are balanced or opposing, weights are distributed between both tasks to establish a balanced, common descent direction.

To optimize model parameters, we integrate the common descent direction $-g$ from MGDA into the Adam optimizer. While Adam typically tracks moments of a stochastic gradient, we utilize it to track the MGDA-derived consensus direction. This approach is justified as $-g$ satisfies the descent condition for all objectives simultaneously. Under the framework of \cite{li2023Adamproof}, convergence is maintained for such descent directions even under relaxed smoothness assumptions.

\subsubsection*{Training procedure}
By combining the proposed loss, the adaptive hyperparameter rule, and MGDA, we establish the comprehensive training process shown in \Cref{alg:mgda}. Denote $\mathcal{L}_1 = \lpoint$ and $\mathcal{L}_2 = \lpi$. At the start of each epoch, the model computes $\text{PICP}^{(\text{train})}$ using the full training set. This step allows the extended log-barrier parameter $r$ to update based on current coverage levels before any weights change. Inside each mini-batch, the model calculates two different forms of loss to guide the learning process.

The first form is the weighted loss, $\mathcal{L}_{\text{weighted}}$, which is used for gradient-based updating $\theta$ with a learning rate $\eta$. This loss is a combination of the point estimation loss $\mathcal{L}_1$ and the PI loss $\mathcal{L}_2$ evaluated on the training set, balanced by the MGDA weights $\gamma_1$ and $\gamma_2$. These weights represent the relative importance of each task gradient calculated from \eqref{eq:closedform_gamma}. By using these $\gamma$ values, the model finds an update direction for model parameters with the Adam optimizer that helps both point and PI forecasts without one task harming the other. The learning rate is managed by a cosine annealing schedule that includes a linear warm-up stage to ensure stable convergence.

The second form is the total validation loss, $\mathcal{L}_{\text{val}}$, which is the sum of $\mathcal{L}_1$ and $\mathcal{L}_2$ on the validation set. This total loss acts as the monitor for the early stopping rule. While the weighted loss changes in every batch to balance gradients, the total validation loss provides a stable metric to check for overfitting. Training stops if $\mathcal{L}_{\text{val}}$ does not improve for a specific number of epochs, known as \texttt{patience}. 

\begin{algorithm}[H]
\caption{Point and prediction interval learning via MGDA}

\begin{algorithmic}[1]

\Require Problem parameters: target probability $p$, Sum-$k$ loss parameter $\lambda < 1$
\Statex \hspace{2.35em} (for solar-context, target probabilities: $p_{\text{day}} = 0.90$, $p_{\text{night}} = 0.15$)
\Statex \hspace{2.35em} Algorithm parameters: \texttt{lr}, \texttt{min\_epoch}, \texttt{max\_epoch}, \texttt{patience}

\State Initialize model parameters $\theta$ \Comment{initial guess for weights and biases}
\State $\mathcal{L}_{\text{val}}^{*} \leftarrow \infty$ \Comment{initialize best validation loss to infinity}
\State $\text{patience counter} \leftarrow 0$

\For{each epoch $e = 1, 2, \ldots$}
    \For{each $k$-step}
    	\State Compute $\text{PICP}_k^{(\text{train})}$  on training set \Comment{\Cref{eq:picp}}
	\State Update log-barrier parameter $r_k$ based on coverage deviations using $p$ \Comment{\Cref{eq:adaptive_r}}
    \EndFor
    \For{each mini-batch $\mathcal{B}$} \Comment{MGDA inner loop}

        \State Infer model outputs: $(\hat{l}_i, \hat{u}_i)$ and $\hat{y}_i$ for all $i \in \mathcal{B}$

        \State Compute point and PI losses: $\mathcal{L}_1$ and $\mathcal{L}_2$ \Comment{\Cref{eq:pointloss_allsteps} and \Cref{eq:piloss_allsteps} } 

        \For{each task $j \in \{1, 2\}$}
            \State $g_j \leftarrow \nabla_{\theta}\, \mathcal{L}_j$ \Comment{compute task gradient}
        \EndFor

        \State Solve MGDA for $(\gamma_1^*, \gamma_2^*)$ \Comment{using \eqref{eq:closedform_gamma}}

        \State $\theta \leftarrow \text{Adam}(\theta, \nabla_\theta \mathcal{L}_\text{weighted} = \gamma_1^* g_1 + \gamma_2^* g_2)$ \Comment{update parameters via Adam optimizer}

    \EndFor

    \State Compute validation loss $\mathcal{L}_{\text{val}} \leftarrow \mathcal{L}_1^{(\text{val})} + \mathcal{L}_2^{(\text{val})}$

    \State \textbf{if} $\mathcal{L}_{\text{val}} < \mathcal{L}_{\text{val}}^{*}$, \textbf{then} $\mathcal{L}_{\text{val}}^{*} \leftarrow \mathcal{L}_{\text{val}}$;\quad $\theta^* \leftarrow \theta$;\quad $\text{patience counter} \leftarrow 0$; \textbf{else} $\text{patience counter} \leftarrow \text{patience counter} + 1$

    \State \textbf{if} $e \geq \texttt{min\_epoch}$ \textbf{and} ($\text{patience counter} \geq \texttt{patience}$ \textbf{or} $e \geq \texttt{max\_epoch}$); \textbf{then} \textbf{break}
\EndFor

\State \Return model with parameters $\theta^*$

\end{algorithmic}
\label{alg:mgda}
\end{algorithm}

\subsubsection*{Example of training and validation loss}

\begin{figure}
\centering
\includegraphics[width=0.95\textwidth]{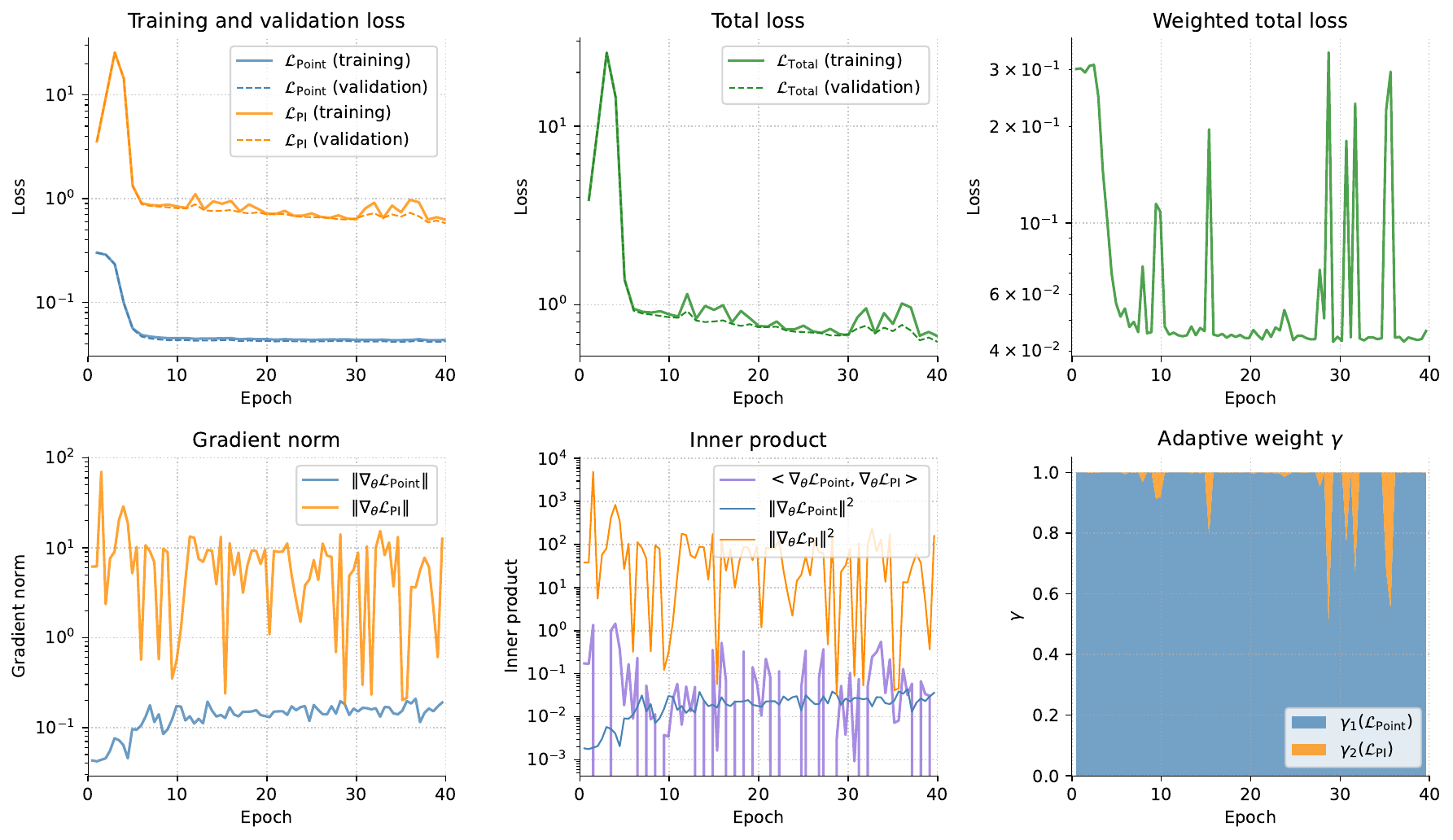}
\caption{MGDA training process: (top) point and PI losses, total loss, and weighted loss; (bottom) gradient norms, inner product between task gradients, and adaptive weights $\gamma_1$, $\gamma_2$.}
\label{fig:mgda_training}
\end{figure}

\Cref{fig:mgda_training} illustrates the training process of the MGDA during initial epochs. The training and validation loss curves show that the PI loss ($\mathcal{L}_\text{PI}$) spikes sharply in the first few epochs. This reflects the large log-barrier penalty when the initial coverage is far below the target. Meanwhile, the point forecast loss ($\lpoint$) stays low and stable throughout the training. Both losses converge within five epochs, and their validation curves track closely with the training curves, indicating that the model does not overfit. The total loss ($\mathcal{L}_{\text{Total}}$) also drops steadily after the initial steps. The weighted total loss reveals occasional sharp spikes at later epochs. These spikes coincide with the adaptive tightening of the barrier parameter ($r$) as the coverage probability approaches its target, and they resolve quickly as the optimizer adjusts.

The behavior of the adaptive weights $\gamma$ is directly governed by the geometric relationship between the loss gradients. \emph{Initial convergence (epochs 0–7):} the gradient of $\lpoint$ is substantially smaller than and contained within the direction of $\lpi$. Geometrically, the projection of $\nabla \lpi$ onto $\nabla \lpoint$ exceeds $\|\nabla \lpoint \|^2$. Consequently, MGDA assigns the entire weight to the point loss ($\gamma_1 = 1$), prioritizing the refinement of the more restrictive task. \emph{Gradient balancing (epochs 25–35):} In the later stages of training, the two gradients become more comparable in orientation. Because neither task clearly contains the other, the optimizer distributes weights between $\gamma_1$ and $\gamma_2$. This distribution represents a balanced trade-off, ensuring a common descent direction that simultaneously satisfies both point accuracy and PI objectives.

MGDA consistently prioritizes the point loss ($\gamma_1 \approx 1$) as a normalizing force. Under standard scalarization, the PI loss would dominate the updates because its gradient is roughly $100$ times larger, effectively obscuring the point loss signal. By assigning the weight to the smaller, more restrictive point loss gradient, MGDA prevents this imbalance. Because the gradients are largely aligned, the PI loss still improves as a free rider, while the model ensures that the sensitive point estimation remains accurately minimized.

\paragraph{Notes on multi-objective framework.} 

The challenge of point and PI forecasting can be framed as a multi-objective optimization problem in several ways. While PICP and PI width are fundamentally conflicting, the relationship between point estimation loss and PI loss is more synergistic. Given that these losses are aggregated over a $k$-step horizon, we explored the following frameworks:
\begin{itemize}
\item Explicit PICP-width splitting: We initially treated PICP and PI width as two independent objectives. However, the model struggled to satisfy both simultaneously, often converging to a trivial solution with near-zero PICP and a small PI width.
\item Composite-loss MGDA: We tried constructing two distinct composite losses, each representing a different weighted combination of PICP and PI width. Loss 1 was designed to penalize PI width more aggressively, while Loss 2 prioritized coverage. We applied the MGDA to dynamically determine the optimal weights for these two composites. Despite this, the resulting PICP remained below target levels.
\item Multi-objective $k$-step losses: One could define $H$ distinct losses, where each loss $k$ combines point and PI estimation for the $k$-th prediction step. However, these losses are highly correlated rather than contradictory, offering little trade-off for a multi-objective framework to exploit. Furthermore, this approach still requires pre-defined weights to balance the point and PI objectives within each step.
\end{itemize}

\section{Data description}
\label{sec:data_description}

\subsection{Dataset}
\begin{figure}
    \centering
    \begin{subfigure}[b]{0.495\textwidth}
        \centering
        \includegraphics[height=8cm,keepaspectratio]{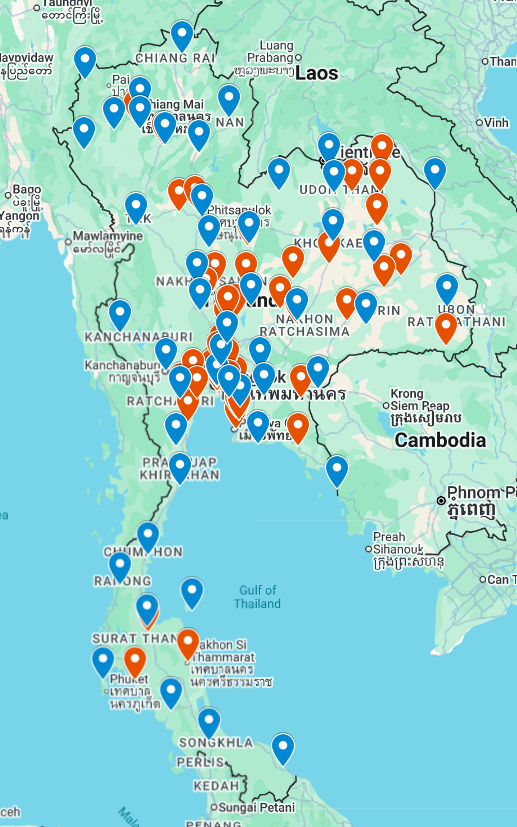}
        \caption{Site location map.}
        \label{fig:location-map}
    \end{subfigure}
    \begin{subfigure}[b]{0.495\textwidth}
        \centering
        \includegraphics[height=8cm,keepaspectratio]{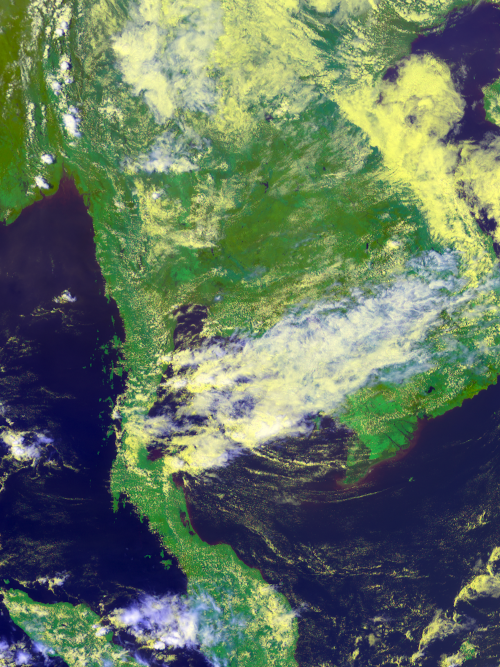}
        \caption{Himawari-8 imagery example.}
        \label{fig:cloud-example}
    \end{subfigure}

    \caption{Overview of geographical data sources: (a) Geographical distribution of the 104 solar measurement stations across Thailand and (b) an example of the Himawari-8 satellite RGB cloud imagery utilized in this study.}
    \label{fig:data-sources-overview}
\end{figure}

\paragraph{Measurements of solar irradiance.} The historical global horizontal irradiance data utilized in this study were collected from 104 measurement stations distributed across Thailand. These stations are operated by the Department of Alternative Energy Development and Efficiency (DeDe), Ministry of Energy, Thailand, as well as private solar farms. All measurements were sampled at a 15-minute resolution. The temporal coverage varies by source: the DeDe dataset spans from January to December 2023, whereas the data provided by the private solar farms cover the period from January 2022 through June 2023. The geographical distribution of these 104 sites is illustrated in \Cref{fig:location-map}. Missing data were addressed via linear interpolation for short gaps lasting less than 6 hours; longer periods of missing observations were retained as NaN values and subsequently excluded during the batch data preparation process.

\paragraph{Re-analysis weather data.} To supplement the ground-based measurements, we incorporated historical reanalysis solar irradiance data ($I_{\text{cams}}$) from the Copernicus Atmosphere Monitoring Service (CAMS). It is crucial to note that this dataset consists of retrospective, reanalyzed historical records rather than real-time monitoring or operational forecasts. CAMS derives the surface solar irradiance by integrating satellite-based cloud parameter observations with detailed aerosol and atmospheric composition data using the Heliosat-4 method \cite{Qu2017}. This reanalysis data provides a spatially and temporally consistent baseline that accounts for historical atmospheric attenuation.

\paragraph{Satellite-derived cloud data.} The cloud cover data were extracted from Himawari-8 satellite RGB imagery \cite{HimawariOptemis}, possessing a spatial resolution of 2 km $\times$ 2 km. These images were captured daily between 06:00 and 19:50 at 10-minute intervals. The cloud index (CI) is computed by normalizing the raw pixel color values: $\text{CI} = \frac{X - \text{LB}}{\text{UB} - \text{LB}}$ where $X$ represents the raw pixel value, and $\text{LB} = 0$ and $\text{UB} = 255$ denote the lower and upper bounds of the 8-bit color spectrum, respectively. In this study, we specifically utilized the cloud mask ($\text{CI}_{CM}$) and the red-channel cloud index ($\text{CI}_{R}$) due to their strong statistical correlation with ground-level solar irradiance attenuation. An example of the Himawari-8 cloud imagery is depicted in \Cref{fig:cloud-example}. To handle missing data, gaps shorter than one hour were interpolated, followed by rolling mean smoothing and temporal resampling to 15-minute resolution.

\begin{figure}
\centering
\includegraphics[width=0.5\textwidth]{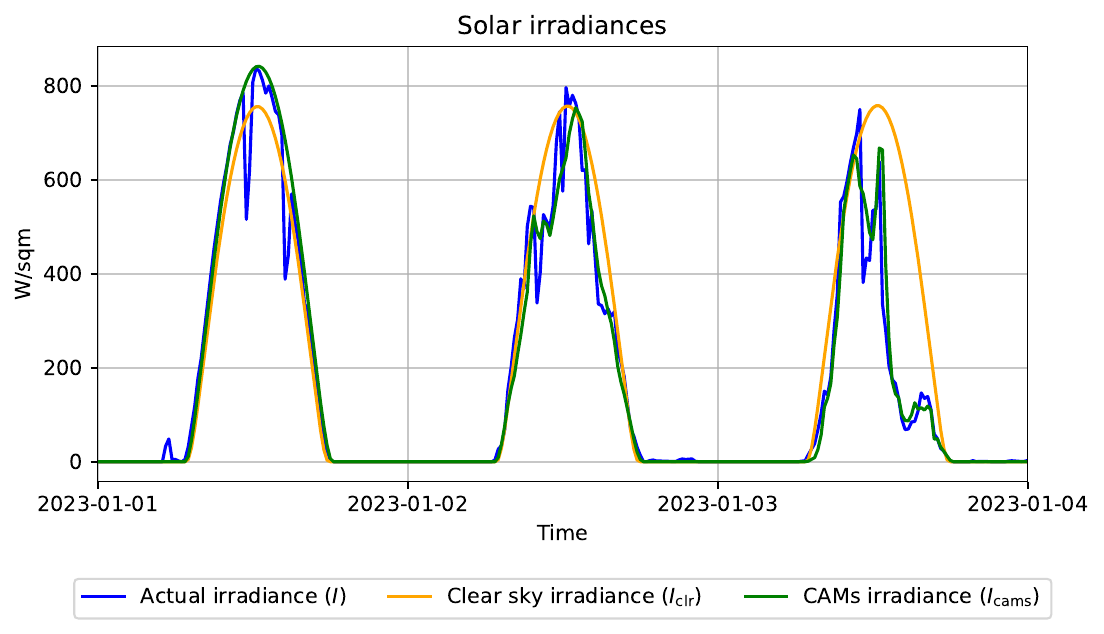}
\caption{A sample time-series of irradiance data.}
\label{fig:irradiance-plot}
\end{figure}

\paragraph{Clear-sky irradiance.} The clear-sky irradiance ($I_{\text{clr}}$) represents the theoretical maximum solar radiation reaching the surface under cloud-free conditions. It is estimated using an improved clear-sky model \cite{Suwanwimolkul2025, EnergyCUEE2024} derived from the original Ineichen–Perez formulation \cite{Ineichen2002}, with the Linke turbidity factor ($T_L$) recalibrated for Thailand to improve local accuracy. The model takes site-specific latitude, longitude, and altitude as inputs to estimate the ideal daily irradiance profile. From this baseline, the clear-sky index is defined as $k = I / I_{\text{clr}}$, which normalizes observed irradiance against the theoretical maximum. This index captures the attenuation of solar radiation due to cloud cover and aerosols, effectively isolating stochastic atmospheric variability from deterministic diurnal and seasonal cycles. \Cref{fig:irradiance-plot} presents a sample time-series profile illustrating the measured irradiance, the clear-sky irradiance, and the CAMS irradiance.

\subsection{Data arrangement}
The experiments considered in this paper involve various model architectures with different data arrangement requirements. Consequently, we employ two distinct data splitting strategies: a stratified shuffling technique and a spatial cross-validation technique. The former is used to evaluate objective functions, while the latter is used to benchmark model architectures.

\paragraph{Regressors.} We incorporate three types of data inputs for the model: auto-regressive features consisting of historical irradiance measurements, exogenous lag regressors obtained from the cloud index, and future regressors including CAMS irradiance, clear-sky irradiance, and an encoded hour index. To preserve periodic continuity between hour 23 and hour 0, the hour index is transformed using a sine-based cyclical encoding: $\sin(\pi h / 24)$, where $h$ denotes the hour index.
 
\paragraph{Data for the experiment of comparing loss functions.}
The experiment in \Cref{subsec:benchmark_objective} compares the effectiveness of loss functions and utilizes the architecture shown in \Cref{fig:pipoint_model} with a 4-hour historical window and a 4-hour forecast horizon. To ensure a uniform data distribution, days are categorized by sky condition (clear, partly cloudy, or cloudy) using the clear-sky index. We then apply daily-block shuffling, where entire 24-hour segments with the same location are randomly permuted. This approach preserves the intraday temporal correlations of each daily time series while randomizing their chronological order.
The dataset of 4,120,032 samples is partitioned into train:validation:test using an 80:10:10 ratio. A stratified split is employed to ensure a balanced representation of sky conditions across all subsets.

\paragraph{Data for the experiment of comparing model architectures.}
In the experiment comparing model architectures (\Cref{subsec:benchmark_models}), specific deep learning implementations require data to be fed in chronological order. Consequently, rather than shuffling temporal blocks, we employ a spatial cross-validation strategy. The dataset of 3,922,496 samples is partitioned entirely by the geographic location of solar measurement stations using an 80:10:10 ratio for train:validation:test. Each subset is curated to ensure spatial diversity, covering all regions of Thailand (North, East, South, and Central). By maintaining the full temporal sequence for each station, every split inherently captures a complete year of seasonal patterns while ensuring the model generalizes across different geographic locations.

\section{Experimental results}
\label{sec:experiments}

This section evaluates the performance of the proposed framework against established baselines and current literature across two key dimensions: the efficacy of the loss function and the integration of  deep learning architectures. Furthermore, we detail the benchmarking framework, the standard metrics employed for evaluation, and the relevant computational considerations. Python implementation is available at \url{https://github.com/energyCUEE/PIPointForecast}.

\subsection{Evaluation metrics}
Let $y$ and $\hat{y}$ denote the ground-truth irradiance and its point estimate, respectively, with $[\hat{l}, \hat{u}]$ representing the prediction interval. Regressors are denoted by $x$. The evaluation metrics are  aggregated over the test set with $N$ samples.

Typical metrics for evaluating point forecast accuracy are MAE, RMSE, and MBE, described by  
\begin{equation}
\text{MAE} = \frac{1}{N} \sum_{i=1}^{N} |y_i - \hat{y}_i|, \quad
\text{RMSE} = \sqrt{\frac{1}{N} \sum_{i=1}^{N} (y_i - \hat{y}_i)^2}, \quad
\text{MBE} = \frac{1}{N} \sum_{i=1}^{N} (y_i - \hat{y}_i).
\label{eq:point_metrics}
\end{equation}

We assess the trade-off between coverage reliability and interval sharpness using several PI metrics. To facilitate comparison across datasets from different studies, these metrics are normalized to ensure scale-independence. To avoid the sensitivities of a min-max range, which can be skewed by outliers, we normalize by the inter-quantile range: $R_Q=q_y(0.95) - q_y(0.05)$.

\begin{itemize}
\item \textbf{Prediction interval coverage probability (PICP):} This metric quantifies interval reliability by measuring the proportion of observed values that successfully fall within the estimated PI. The PICP is defined as:
\begin{equation}
\text{PICP} = \frac{1}{N}\sum_{i = 1}^N \mathbf{1}\{\hat{l}_i \le y_i \le \hat{u}_i\},
\end{equation}
where $\mathbf{1}\{E\}$ is a counting function equal to $1$ if $E$ is true, and $0$ otherwise. The PICP is expected to approach the nominal confidence level $p$ specified in the training process.

\item \textbf{Prediction interval normalized average width (PINAW):} This metric evaluates the sharpness of PI where lower PINAW values signify narrower, sharper intervals, and are highly preferable provided the target PICP is achieved. 
\begin{equation}
    \text{PINAW} = \frac{1}{NR_Q}\sum_{i=1}^N (\hat{u}_i - \hat{l}_i),
\end{equation}
Without data-scale normalization, the mean prediction interval width (MPIW) is also used in literature. 
\item \textbf{Prediction interval normalized average large width (PINALW):} While average width is a standard metric, it may not fully capture the critical characteristics of PIs, particularly since large width samples can disproportionately drive the conservatism in robust design. To address this, PINALW \cite{Amnuaypongsa2025} computes the average of the \emph{$K$-largest} PI widths, specifically targeting samples that exceed the $\tau$-quantile of the width distribution.
\begin{equation}
    \text{PINALW}(\tau) = \frac{1}{KR_Q}\sum_{i=1}^K|w|_{[i]},
\end{equation}
where $|w|_{[i]}$ denotes the $i^\text{th}$ largest absolute PI width, \ie, $|w|_{[1]} \geq |w|_{[2]} \geq \cdots \geq |w|_{[N]}$, and $K = \lfloor(1-\tau)N\rfloor$ is the total number of samples exceeding the $\tau$-quantile threshold. For example, in our evaluation, $\tau$ is set to $0.5$, giving the average of the largest $50\%$ of all widths to represent the large width samples.

\item \textbf{Winkler score \cite{Winkler1972}:} The Winkler score offers a comprehensive evaluation by jointly capturing sharpness and reliability. It penalizes wide intervals and imposes an additional, more severe penalty when an observation falls outside the constructed bounds. For a PI with nominal probability $p$, the normalized Winkler score is defined as:

\begin{equation}
    \text{Winkler} = \frac{1}{NR} \sum_{i=1}^N\left(|\hat{u}_i - \hat{l}_i| + \frac{2}{1-p}\left[(\hat{l}_i-y_i)\mathbf{1}\{y_i < \hat{l}_i\} + (y_i-\hat{u}_i)\mathbf{1}\{y_i > \hat{u}_i\}\right]\right).
\end{equation}
A lower Winkler score indicates superior model performance, implying that the constructed lower and upper bounds align tightly with the theoretical quantiles of $(1-p)/2$ and $(1+p)/2$.
\end{itemize}

While other probabilistic metrics exist, such as CRPS \cite{Yang2020}, they are inapplicable to direct PI forecasts. Additionally, although CWC \cite{Khosravi2011} combines PICP and width into a single score, it requires a hyperparameter to balance these terms; we prefer to evaluate them individually to maintain clarity.

\paragraph{Evaluation setting.} We evaluate solar irradiance forecasting performance exclusively during \emph{daytime} hours (06:00–18:00) to ensure the metrics reflect actual modeling challenges rather than trivial nighttime zeros. Since nighttime data accounts for half of the dataset, its inclusion would artificially lower the MAE by a factor of $1/2$ and the RMSE by $1/\sqrt{2}$, leading to an overly optimistic assessment of model accuracy \cite{Solartalk2025}. The MAE, RMSE, and MBE are reported in $\wm$ for comparison consistency with other solar energy studies, while the PI metrics are scale-independent and unitless.

\subsection{Loss comparison}
\label{subsec:benchmark_objective}
This section compares the efficacy of the proposed loss function against several established methods from prior research. To ensure a fair comparison, the underlying model architecture is kept consistent with our proposed design shown in \Cref{fig:pipoint_model} across all experiments. Specifically, for the QR, QD+, and EMQ methods, the models are configured to directly output three distinct values: the point forecast $\hat{y}$, the upper bound $\hat{u}$, and the lower bound $\hat{l}$. Conversely, the IPIV method strictly utilizes its proposed architectural design, which constructs the point forecast via a convex combination of the bounds. While all loss functions, $\hat{y}, \hat{l},\hat{u}$, PICP and widths are functions of the model parameters $\theta$ and regressors $x$, we omit these arguments hereafter to simplify the notation. The following benchmark methods have been adapted for our evaluation:

\begin{enumerate}
\item \textbf{Quantile regression (QR)}:
Quantile regression directly estimates target quantiles by minimizing the pinball function defined as $\rho_\tau(r) = \max\{\tau r, (\tau-1)r\}$ for a given quantile $\tau$. To produce a PI that covers the target probability of $1-\alpha$, we use the pinball loss consisting of two quantile terms: $\mathcal{L}_{\text{QR}} = \frac{1}{N} \sum_{i=1}^N  \rho_{\alpha/2}\big(y_i - \hat{l}) + \rho_{1-\alpha/2}\big(y_i - \hat{u} )$.

\item \textbf{Quality-driven loss (QD+) \cite{Pearce2018}}:
Building upon the standard quality-driven loss, QD+ incorporates point forecasting and boundary constraints with the loss function consisting of four terms: the first two represent the original QD loss that is served for width penalty and PICP, while the latter two terms signify the MSE loss and penalty on the forcing the $\hat{y}$ to lie within the PI. 
\begin{equation} 
 \mathcal{L}_{\text{QD+}} = (1-\lambda_{1})(1-\lambda_{2})\text{MPIW}_{\text{capt.}} + \lambda_{1}(1-\lambda_{2})\max\big(0, (1-\alpha) - \text{PICP}\big)^{2} + \lambda_{2}\mathcal{L}_{\text{MSE}} + \epsilon \mathcal{L}_{\text{Penalty}}.
\label{eq:qdplus}
\end{equation}
where the $\text{MPIW}_{\text{capt.}}$ is the mean PI width penalty that measures only the samples that $y$ lies within the PI, and $ \mathcal{L}_{\text{Penalty}} = \frac{1}{N}\sum_{i=1}^{N} \big[ \max(0, \hat{l}_{i} - \hat{y}_{i}) + \max(0, \hat{y}_{i} - \hat{u}_{i}) \big] $ penalizes violations of $\hat{y} \in [\hat{l},\hat{u}]$. The hyperparameter $\lambda_{1} \in (0, 1)$ balances the PI width and PICP, while $\lambda_{2} \in (0, 1)$ determines the weight of the point forecast error ($\mathcal{L}_{\text{MSE}}$) relative to the PI metrics. The hyperparameter $\epsilon$ scales the penalty for constraint violations. A notable drawback of this approach is the necessity to tune $\lambda_{1} $, $ \lambda_{2}$, and $\epsilon$ set by the user's preference. Furthermore, the soft penalization does not strictly guarantee that the point forecast $\hat{y}$ will remain bounded within the interval bounds.

\item \textbf{Integrated prediction interval and value predictions (IPIV) \cite{Simhayev2022}}: The method produces three outputs: $\hat{y},\hat{l},v \in (0,1)$ and constructs the point forecast via a convex combination: $\hat{y} = v \hat{u} + (1-v) \hat{l}$ which helps guarantee that $\hat{y}$ always lies within the PI, while positioning $\hat{y}$ asymmetrically. The loss is a weighted sum of the PI and point forecast objectives: $\mathcal{L}_{\text{IPIV}} = \beta \lpi + (1-\beta) \lpoint$ with a weight $\beta$ (default 0.5) where 
\begin{equation} 
\lpi = \text{MPIW}_{\text{capt.}} + \sqrt{N} \lambda \max\big(0, (1 - \alpha) - \text{PICP}\big)^{2}, \;\; \lpoint = \frac{1}{N}\sum_{i=1}^{N} \ell\big(v_{i} \hat{u}_{i} + (1-v_{i})\hat{l}_{i}, y_{i}\big)
\label{eq:ipiv}
\end{equation}
The hyperparameter $\lambda$ governs the balance between PICP and interval width. The point forecast loss utilizes a regression loss $\ell$. To better capture epistemic model uncertainty, IPIV further incorporates an ensemble strategy across multiple networks and a $z$-score calibration step.

\item \textbf{Enhanced multi-quantile loss (EMQ) \cite{Saeed2025a}}: The method addresses three main challenges: poor estimation of extreme tail quantiles, quantile crossing, and suboptimal balance between PI width and the target $1-\alpha$ coverage: 
\begin{equation} 
\mathcal{L}_{\text{EMQ}} = \mathcal{L}_{\text{adj. pinball}} + \lambda \mathcal{L}_{\text{order}} + \mathcal{L}_{\text{covarage}} + \text{MPIW}_{\text{scaled}}.
\label{eq:emq}
\end{equation}
To improve tail estimation, $\mathcal{L}_{\text{adj. pinball}}$ aggregates pinball losses across a $3$-neighbor set around $\alpha/2$ and $1-\alpha/2$ quantiles, plus the median, reducing sensitivity to any single extreme quantile estimate. The $\mathcal{L}_{\text{order}}$ term  penalizes any pair of adjacent quantiles violations via an $\exp( \max(0,\cdot))$ function. The term $\mathcal{L}_{\text{covarage}}$ term penalizes samples $y_i$ falling outside the PI, while $\text{MPIW}_{\text{scaled}}$ weights each width sample proportionally to the coverage deficiency (smaller weight) when the nominal level $1-\alpha$ is not met. 
\end{enumerate}

\begin{figure}
\centering
\includegraphics[width=\textwidth]{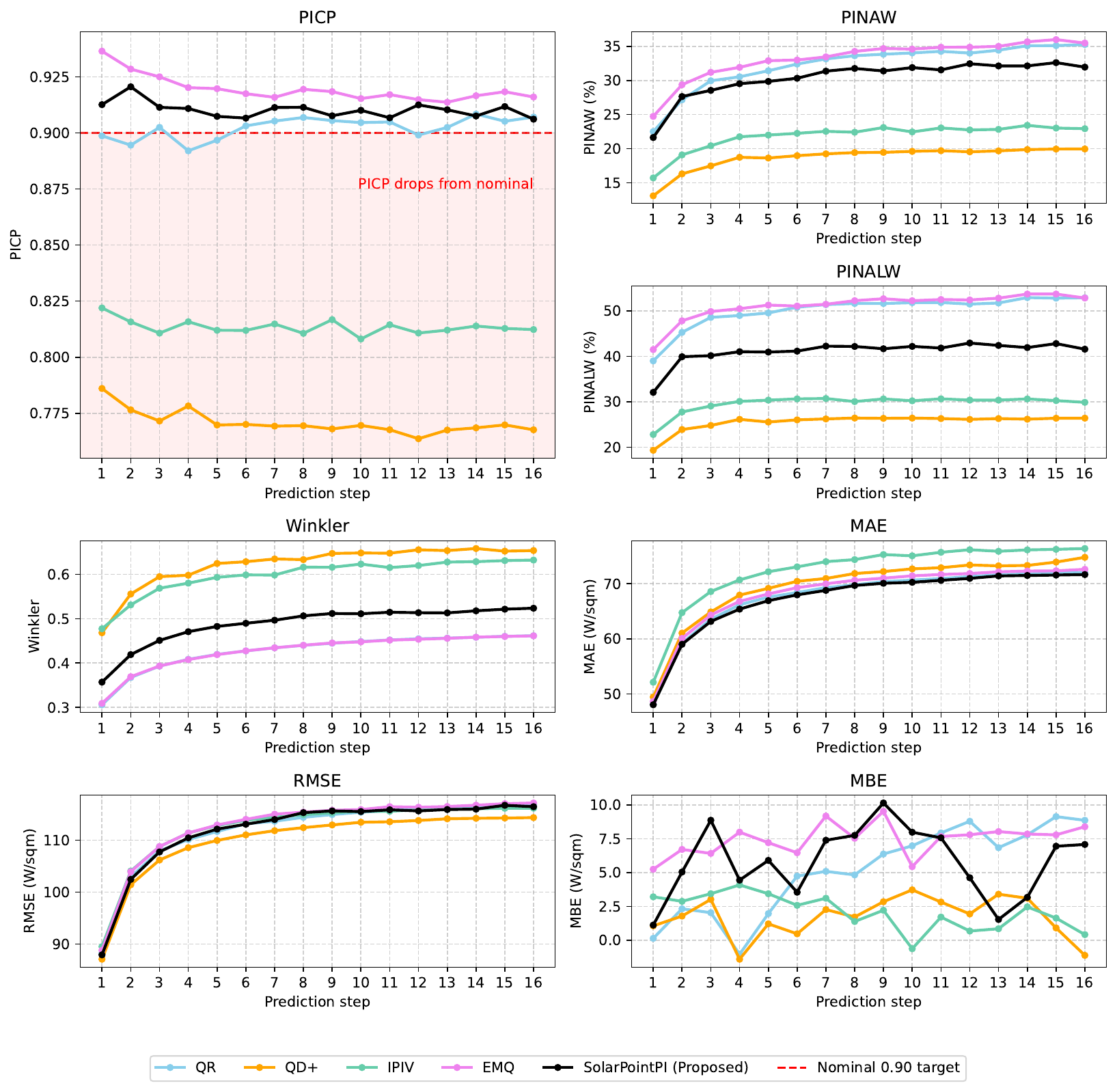}
\caption{Overall performance metrics in objective function comparison.}
\label{fig:compare_loss_metric}
\end{figure}

The performance results are summarized in \Cref{tab:compare_loss} and visually depicted in \Cref{fig:compare_loss_metric}. The evaluation is restricted to daytime data to avoid an overoptimistic evaluation as covering trivial zero-irradiance at nighttime is effortless.

\paragraph{Prediction interval coverage probability (PICP):} \Cref{tab:compare_loss} clearly illustrates a major strength of the SolarPointPI loss that consistently maintains the PICP above the 0.9 nominal level across every time step. Conversely, \Cref{fig:compare_loss_metric} demonstrates how other loss functions suffer from severe coverage degradation. QD+ and IPIV systematically fail to reach the nominal level (yielding values as low as 0.768).  Furthermore, while QR initially meets the requirement, its trajectory drops below the nominal line as the forecasting step extends. Meanwhile, EMQ maintains coverage but tends to over-cover (\eg, 0.936 at the 15-minute horizon), which is compromised to unnecessarily wide intervals.

We observed that EMQ and IPIV achieve the target PICP on the test set if nighttime data \emph{were included} in a separate evaluation setting. Conversely, QD+ and QR fail to reach the target coverage even with nighttime data; for QD+, this is likely due to its high sensitivity to hyperparameters, which were set according to the original study.

\paragraph{Interval sharpness (PINAW and PINALW):} Among the valid models that pass PICP criterion, the SolarPointPI loss exhibits the lowest PINAW and PINALW score across \emph{all} prediction steps; see \Cref{fig:compare_loss_metric}. Notably, maintaining a significant lower PINALW at longer horizons is non-trivial, as interval widths naturally expand with increasing lead times. This performance is driven by the efficiency of the Sum-$k$ width penalty in the proposed PI loss. In contrast, benchmark methods like EMQ and QR (where valid) suffer from significantly inflated PINAW and PINALW scores (reaching PINALW of 0.528 at the 4-hour horizon). Although QD+ and IPIV report low PINAW in \Cref{tab:compare_loss}, this is solely because they fail the PICP requirement. When unnormalizing PINAW to the solar irradiance unit $\wm$, the average widths range in 196-296 $\wm$ across from 15-minute to 4-hour horizons, and the average large widths range in 291-390 $\wm$.

\paragraph{Winkler score:} While the Winkler score provides a convenient single-index summary of both interval sharpness and quantile matching, it can be deceptive if viewed in isolation. As shown in \Cref{tab:compare_loss}, our SolarPointPI maintains a competitive Winkler score, even though EMQ and QR (where valid) achieve better values. However, they suffer from heavily inflated absolute widths to achieve this.

\paragraph{Point forecast accuracy (MAE, RMSE, and MBE):} The SolarPointPI achieves the lowest MAE across all prediction steps and the second lowest RMSE to QD+ (which is not valid by PICP criterion), while MBE can be relatively high. Conversely, IPIV and QD+ struggle with high residual errors, highlighting the difficulty of balancing point and interval objectives. Meanwhile, valid models like EMQ exhibit visibly increasing point errors as the prediction step extends. 

In conclusion, the SolarPointPI loss not only provides rigorous uncertainty quantification that maintains target coverage while reducing PI width, but also ensures the point forecast remains acceptably accurate. These outcomes stem from three components: the log-barrier penalty for reliability, the Sum-$k$ width penalty for sharpness, and the MGDA algorithm for balancing point and PI losses.

\begin{figure}
\centering
\includegraphics[width=0.7\textwidth]{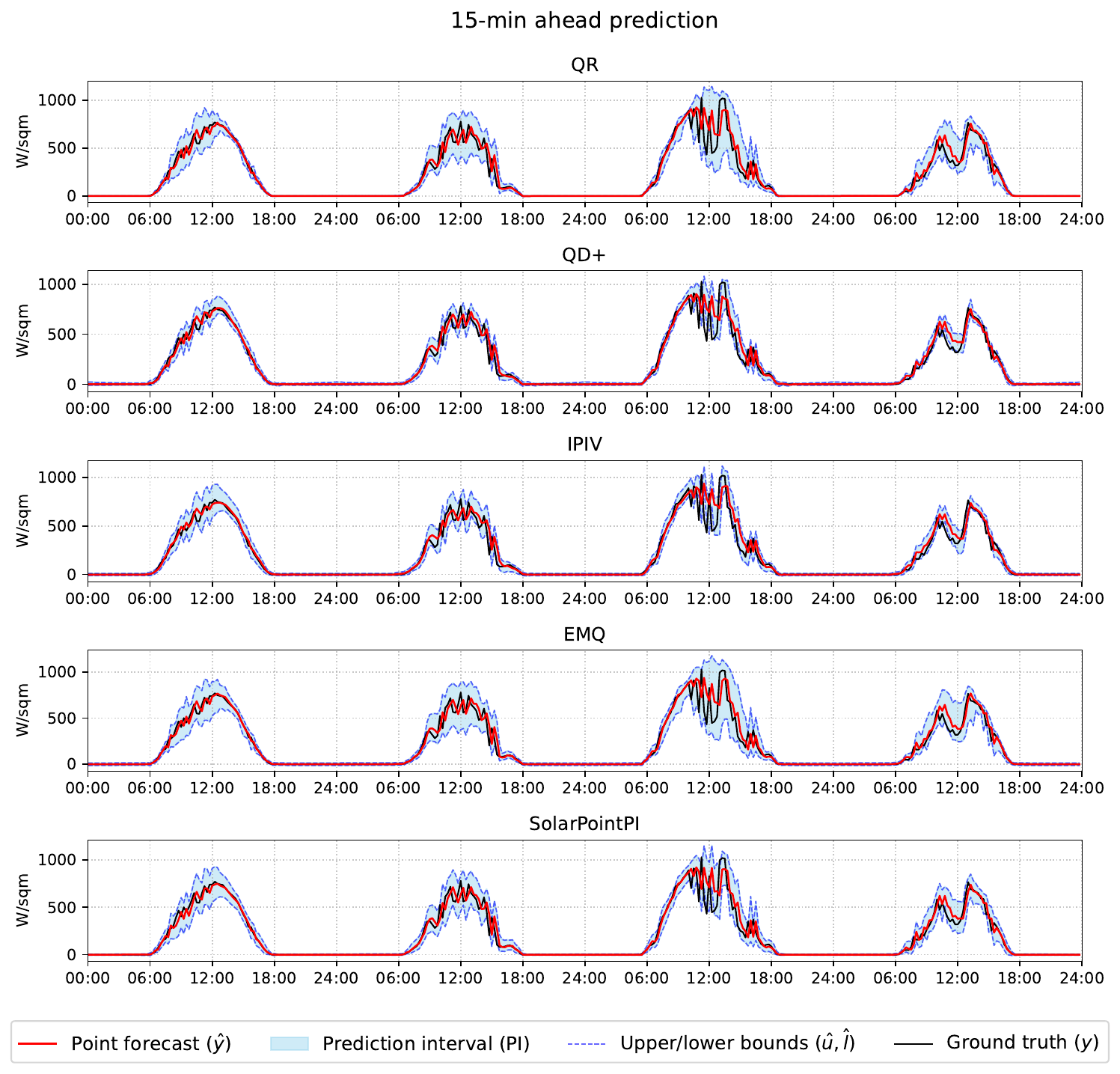} 
\includegraphics[width=0.7\textwidth]{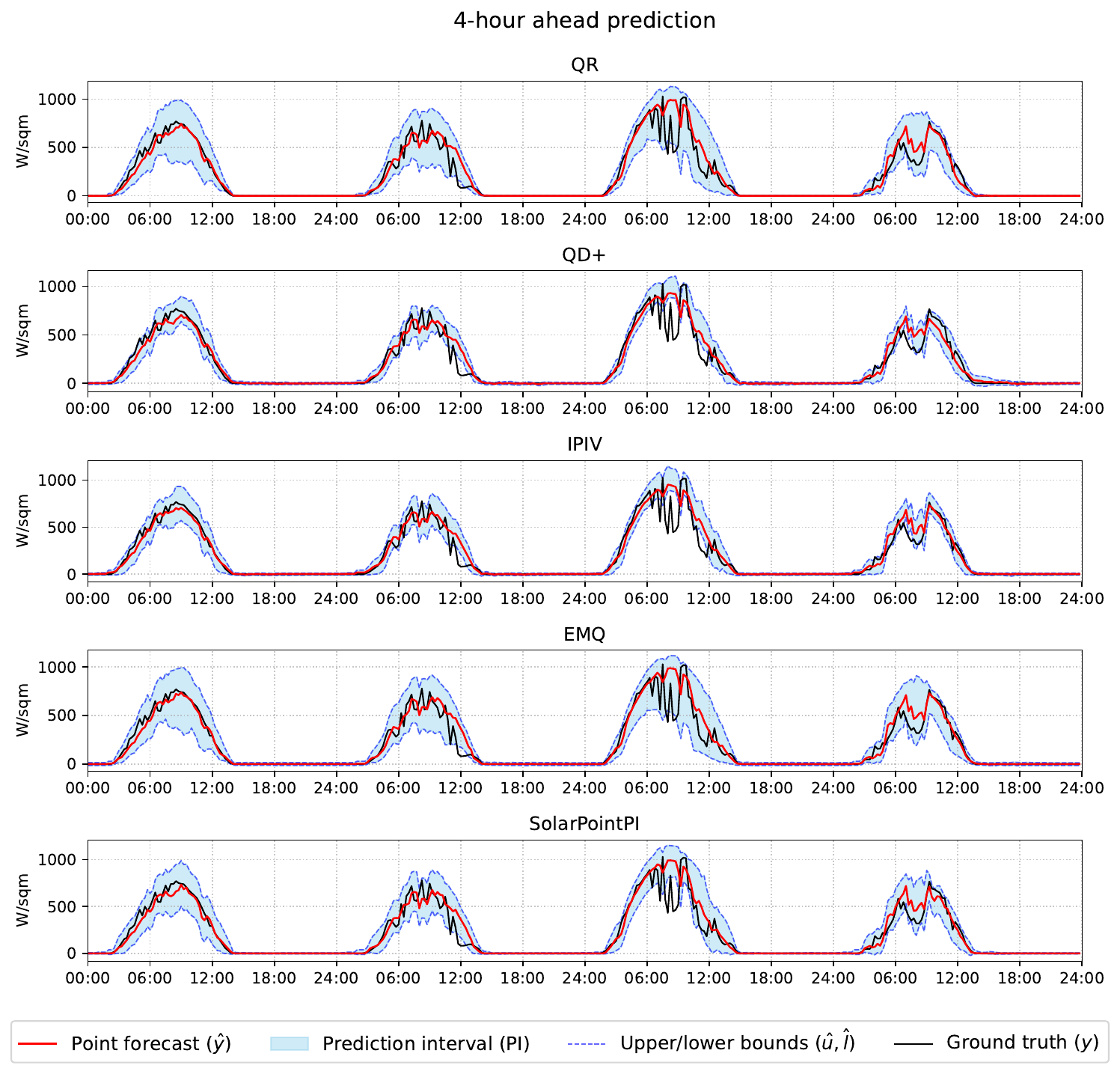}
\caption{Solar irradiance forecasts (W/sqm) comparing loss functions at 15 minute (top) and 4-hour (bottom) horizons under cloudy-sky conditions.}
\label{fig:compare_loss_ts}
\end{figure}

\paragraph{Forecasting result time series:} \Cref{fig:compare_loss_ts} presents the time series plots for solar irradiance forecasts at 15-minute and 4-hour horizons under cloudy sky conditions. The results reveal a clear trade-off: while QR and EMQ maintain coverage across all horizons, they do so through excessively wide PIs. Conversely, QD+, IPIV, and SolarPointPI generate tighter PIs that closely track $y$; however, QD+ and IPIV fail to meet PICP requirements. During high volatility on the third day, SolarPointPI successfully captures the data with narrow intervals at the 15-minute horizon, though coverage slightly degrades at the 4-hour mark. Overall, SolarPointPI provides an optimal balance between reliability and interval sharpness.

\begin{table}
    \centering
    \caption{Performance of \textbf{benchmarking loss functions} evaluated on daytime data (06:00 - 18:00) in the test set. PICP $< 0.9$ is highlighted in \textcolor{red}{red}. Best metrics are bolded (restricted to models with PICP $\geq 0.9$). Optimal scores are the value closest to zero for MBE, and the minimum value for all other metrics. PINAW and PINALW are in \%, while MAE, RMSE, and MBE are in $\wm$. The proposed method is \textcolor{teal}{\bf SolarPointPI}.}
\small
\begin{tabular}{lccccccc}
\toprule
\textbf{Objective} & \multicolumn{7}{c}{\textbf{Metrics}} \\
\cmidrule(lr){2-8}
& \textbf{PICP} & \textbf{PINAW} & \textbf{PINALW} & \textbf{Winkler} & \textbf{MAE} & \textbf{RMSE}& \textbf{MBE} \\
\toprule

\multicolumn{8}{c}{\textbf{15-minute ahead}} \\
\midrule
QR & \color{red} 0.899 & 22.50 & 39.01 & 0.306 & 48.11 & 88.19 & 0.15 \\
QD+ & \color{red} 0.786 & 13.08 & 19.35 & 0.468 & 49.45 & 87.06 & 1.07 \\
IPIV & \color{red} 0.822 & 15.71 & 22.82 & 0.477 & 52.14 & 89.48 & 3.21 \\
EMQ & 0.936 & 24.73 & 41.50 & \bf 0.309 & 48.65 & 88.78 & 5.25 \\
\textcolor{teal}{\bf SolarPointPI} & 0.913 & \bf 21.63 & \textbf{32.09} & 0.357 & \textbf{48.05} &  \textbf{87.90} & \textbf{1.13} \\
\midrule

\multicolumn{8}{c}{\textbf{1-hour ahead}} \\
\midrule
QR & \color{red} 0.892 & 30.53 & 48.99 & 0.408 & 66.18 & 110.14 & -1.03 \\
QD+ & \color{red} 0.778 & 18.73 & 26.15 & 0.598 & 67.94 & 108.57 & -1.39 \\
IPIV & \color{red} 0.816 & 21.74 & 30.10 & 0.580 & 70.69 & 111.40 & 4.09 \\
EMQ & 0.920 & 31.92 & 50.48 & \textbf{0.408} & 66.80 & 111.41 & 7.99 \\
\textcolor{teal}{\bf SolarPointPI}  & 0.911 & \bf 29.52 & \bf 41.03 & 0.471 & \textbf{65.39} & \bf 110.46 & \bf 4.45 \\
\midrule

\multicolumn{8}{c}{\textbf{2-hour ahead}} \\
\midrule
QR & 0.907 & 33.62 & 51.64 & \bf 0.440 & 69.79 & \bf 114.42 & \bf 4.83 \\
QD+ & \color{red} 0.769 & 19.42 & 26.44 & 0.633 & 71.85 & 112.43 & 1.73 \\
IPIV & \color{red} 0.811 & 22.40 & 30.06 & 0.616 & 74.37 & 114.93 & 1.39 \\
EMQ & 0.919 & 34.25 & 52.25 & \textbf{0.440} & 70.63 & 115.42 & 7.54 \\
\textcolor{teal}{\bf SolarPointPI} & 0.911 & \bf 31.76 & \bf 42.18 & 0.506 & \textbf{69.68} & 115.33 & 7.75 \\
\midrule

\multicolumn{8}{c}{\textbf{4-hour ahead}} \\
\midrule
QR & 0.907 & 35.27 & 52.89 & 0.462 & 72.01 & 116.84 & 8.87 \\
QD+ & \color{red} 0.768 & 19.95 & 26.42 & 0.654 & 74.79 & 114.37 & -1.11 \\
IPIV & \color{red} 0.812 & 22.92 & 29.89 & 0.632 & 76.38 & 116.18 & 0.42 \\
EMQ & 0.916 & 35.48 & 52.84 & \textbf{0.461} & 72.61 & 117.19 & 8.39 \\
\textcolor{teal}{\bf SolarPointPI}  & 0.906 & \bf 31.95 & \bf 41.59 & 0.524 & \textbf{71.66} & \bf 116.47 & \bf 7.08 \\
\bottomrule
\end{tabular}
\label{tab:compare_loss}
\end{table}

\clearpage
\subsection{Model comparison} 
\label{subsec:benchmark_models}

We compare our proposed architecture against state-of-the-art models. These include a zero-shot foundation model Chronos 2.0, LightGBM, LSTM and Transformer models configured in an encoder-decoder architecture. To ensure consistency across all evaluations, every model is configured to release a point forecast and PI with a confidence level of $0.9$. Regarding the input window, all models except LightGBM use a 2-day lag, while LightGBM uses a 4-hour lag; both settings target a 4-hour forecasting horizon. All models except the Chronos and LightGBM are trained according to the proposed loss and \Cref{alg:mgda}.

\begin{figure}
\centering
\begin{subfigure}{0.8\textwidth}
\centering
\includegraphics[width=\textwidth]{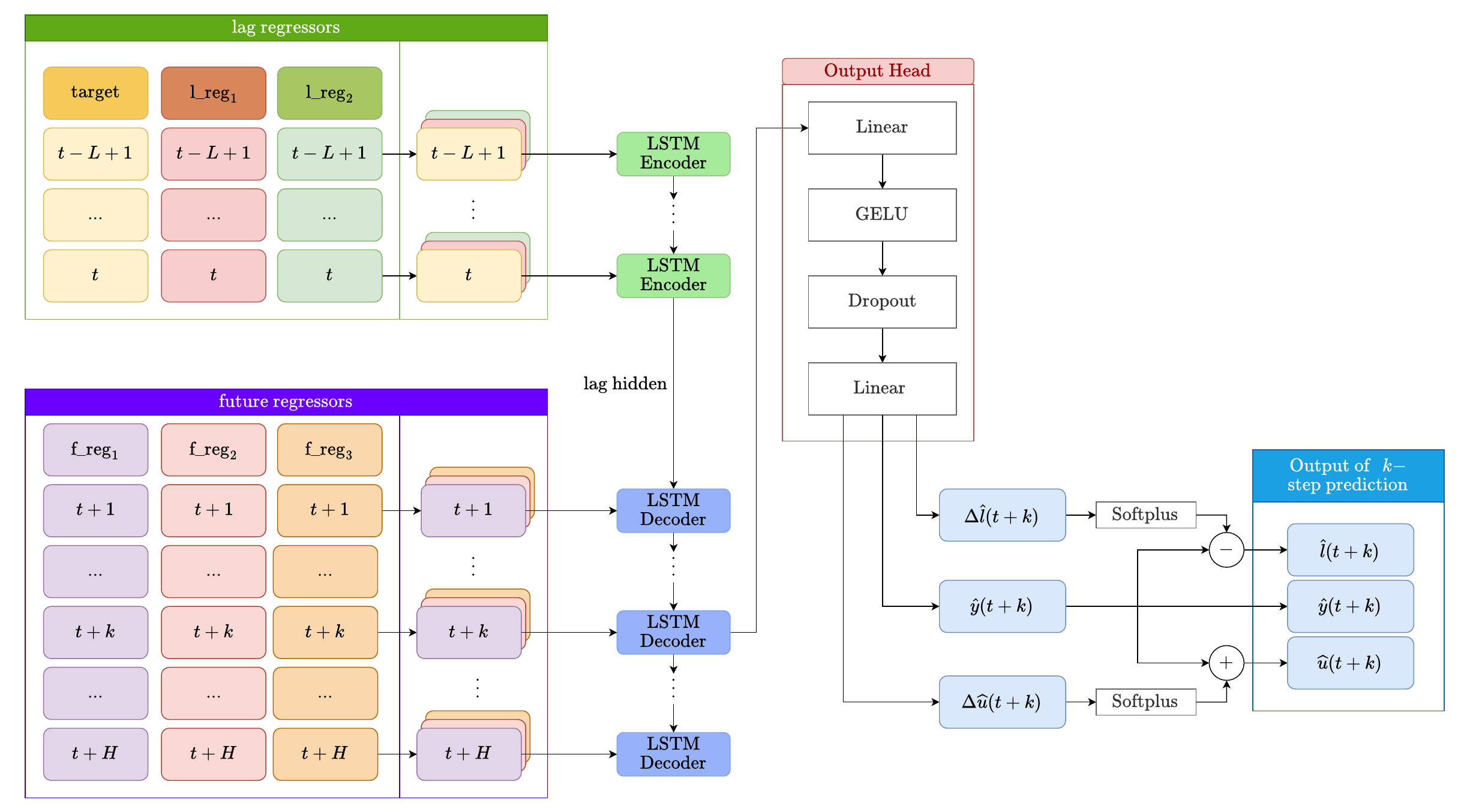}
\caption{LSTM.}
\label{fig:model_architecture_LSTM}
\end{subfigure}

\begin{subfigure}{0.8\textwidth}
\centering
\includegraphics[width=\textwidth]{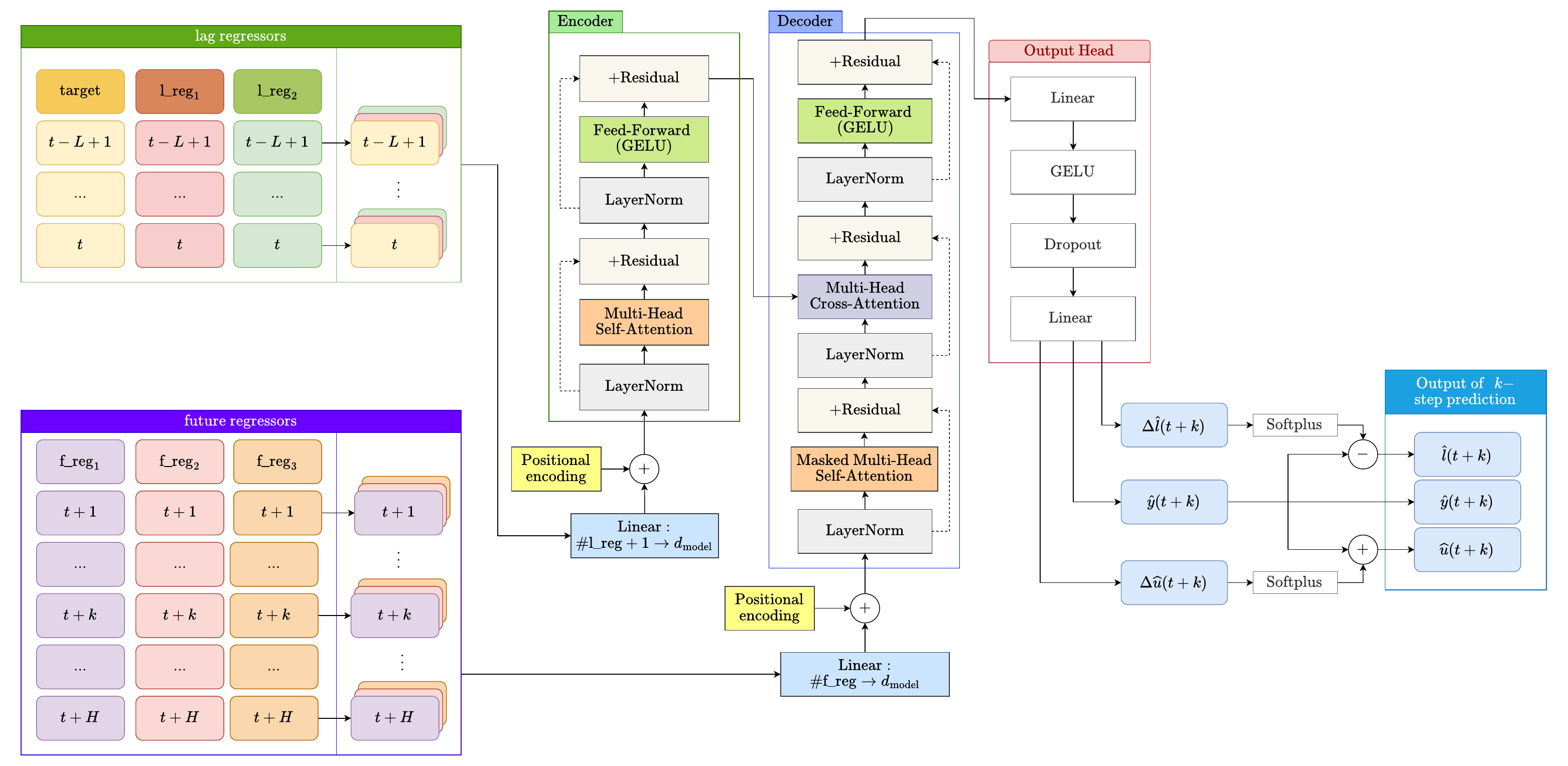}
\caption{Transformer.}
\label{fig:model_architecture_transformer}
\end{subfigure}

\begin{subfigure}{0.8\textwidth}
\centering
\includegraphics[width=\textwidth]{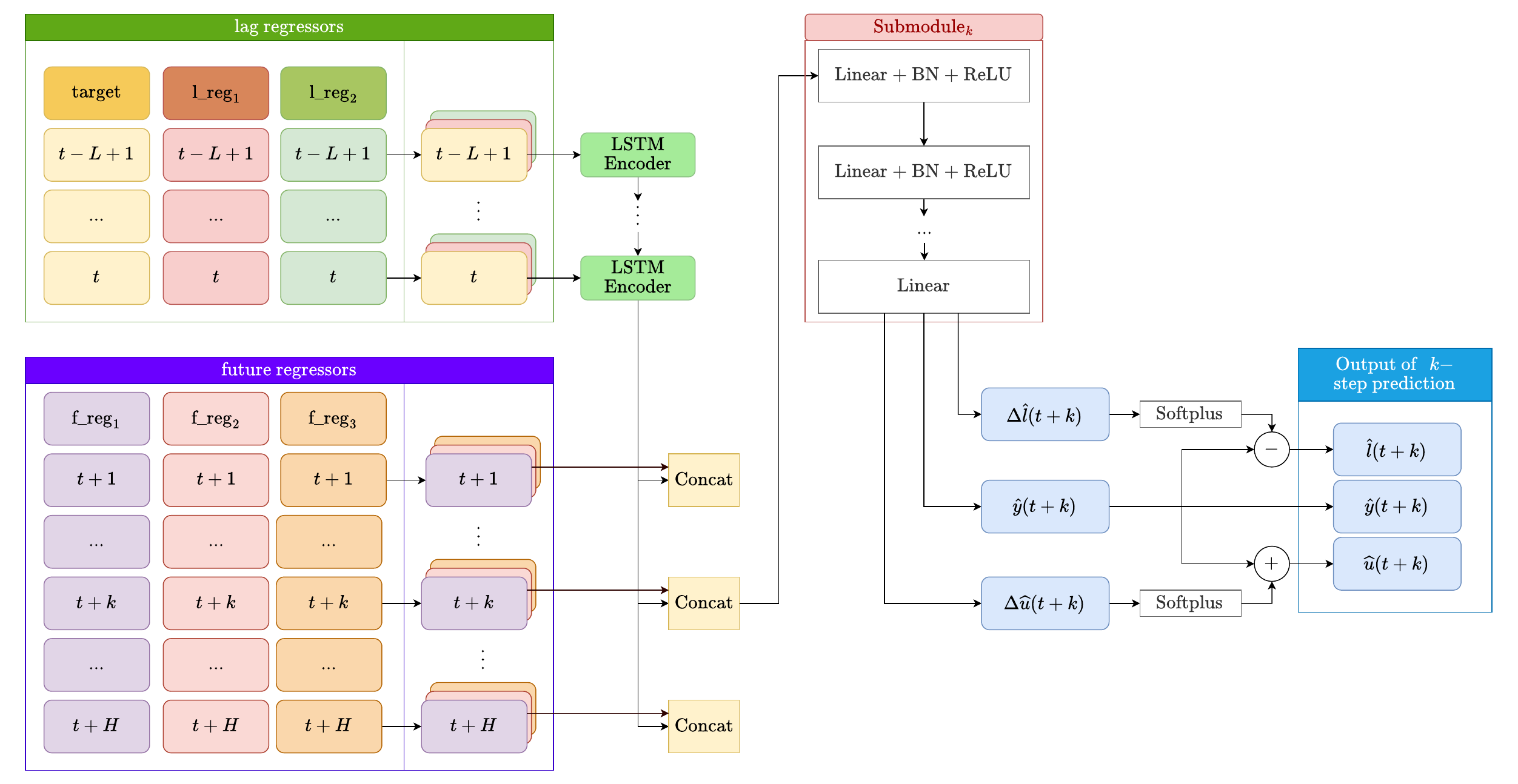}
\caption{SolarPointPI.}
\label{fig:model_architecture_SolarPointPI}
\end{subfigure}
\caption{Benchmarking model architectures.}
\end{figure}

\begin{enumerate}
\item \textbf{Chronos (Zero-shot)} \cite{Ansari2025}: Chronos-2 is a pretrained foundation model based on an encoder-only transformer architecture, designed for zero-shot generalization across universal time series forecasting tasks. It introduces a group attention mechanism that facilitates in-context learning by sharing information across related time series, allowing the model to handle univariate, multivariate, and covariate-informed tasks without the need for task-specific training.
\item \textbf{LightGBM}: LGBM employs a leaf-wise growth strategy and built-in feature selection for modeling efficiency. To generate multi-step forecasts, we train 16 independent models, each optimized via pinball loss for the median (the 0.5-quantile) and PI bounds (the 0.05- and 0.95-quantiles). Hyperparameter tuning (learning rate, depth, and leaf constraints) is performed on the 1-step and 9-step models with the resulting parameters shared across the 1-8 and 9-16 horizons, respectively.
\item \textbf{LSTM}: A recurrent neural network architecture designed for sequence-to-sequence modeling. As shown in \Cref{fig:model_architecture_LSTM}, the LSTM encoder processes historical inputs (lag regressors) and passes its final hidden state to an LSTM decoder. The decoder recursively incorporates future regressors to generate step-wise hidden states. These states are passed through an output head consisting of linear, GELU (Gaussian Error Linear Unit), and dropout layers. The final layer outputs a central point estimate alongside softplus-activated width adjustments ($\Delta \hat{l}_i, \Delta \hat{u}_i$) to strictly formulate the lower and upper interval bounds. 
\item \textbf{Transformer}: An architecture utilizing attention mechanisms to capture long-range dependencies. As shown in \Cref{fig:model_architecture_transformer}, the encoder processes positionally-encoded lag regressors via multi-head self-attention. The decoder utilizes multi-head cross-attention to integrate the encoder's continuous representations with positionally-encoded future regressors. The aggregated features are routed through the output head (same as the one used in LSTM) to yield $\hat{y}, \Delta \hat{l}_i, \Delta \hat{u}_i$.
\item \textbf{Proposed SolarPointPI}: Our primary proposed LSTM architecture. As shown in \Cref{fig:model_architecture_SolarPointPI}, the core LSTM network processes the sequence of lag regressors, and the final state is normalized and activated via batch normalization (BN) and ReLU layers. Rather than using a sequential decoder, this encoded representation is concatenated step-wise with the respective future regressors. The combined vector is processed by step-specific independent submodels (consisting of stacked Linear, BN, and ReLU blocks) acting as specialized heads. Finally, these submodels output the point forecast and boundary adjustments ($\Delta \hat{l}_i, \Delta \hat{u}_i$) to construct the prediction interval.

\item \textbf{Chronos as feature generators:} We augment the Chronos foundation model as a feature extractor. Chronos processes the lag regressors to generate auxiliary future representations as the 0.05, 0.5, and 0.95 quantile outputs. These extracted features are then concatenated with the future regressors before being fed into each of the LSTM, Transformer, and SolarPointPI models.
    \begin{itemize}
    \item Chronos + LSTM: Chronos outputs are applied into the decoder of the LSTM.
    \item Chronos + Transformer: The Chronos outputs are fed into the decoder module of the Transformer.
    \item Chronos + SolarPointPI: The Chronos outputs are concatenated with the outputs of the shared model $\mathcal{M}_c$ and future regressors before being fed into the $k$-step submodels.
    \end{itemize}
\end{enumerate}

\begin{table}[ht]
\centering
\small
\caption{Model and algorithm hyperparameters for model comparison experiment.}
\begin{tabular}{ll} \hline
\toprule
\bf Model & \bf Hyperparameters  \\
\midrule
LGBM & Submodel 1-8: learning rate: 8.8e-3, num leaves: 454, max depth: 14  \\ 
& min data in leaf: 477, feature fraction: 0.63, bagging fraction: 0.58, min gain to split: 0.77 \\
& Submodel 9-16: learning rate: 1.1e-2, num leaves: 456, max depth: 14  \\ 
& min data in leaf: 11, feature fraction: 0.98, bagging fraction: 0.76, min gain to split: 0.86 \\
LSTM & hidden size: 64, num encoder/decoder layer: 2,2, batch size: 32678, learning rate: 1e-4, optimizer: AdamW \\
Transformer & d model: 64, num encoder/decoder layer: 3,3, num heads: 8, feed forward dim: 128, \\
& batch size: 16384 (Gradient accumulation), learning rate: 3e-4, optimizer: AdamW\\
SolarPointPI & num lstm cell: 1, lstm hidden size: 70, submodel neuron: [100,100], \\
& batch size: 32678, learning rate: 3e-4, optimizer: AdamW \\ 
\bottomrule
\end{tabular}
\end{table}

\begin{figure}
\centering
\includegraphics[width=0.8\textwidth]{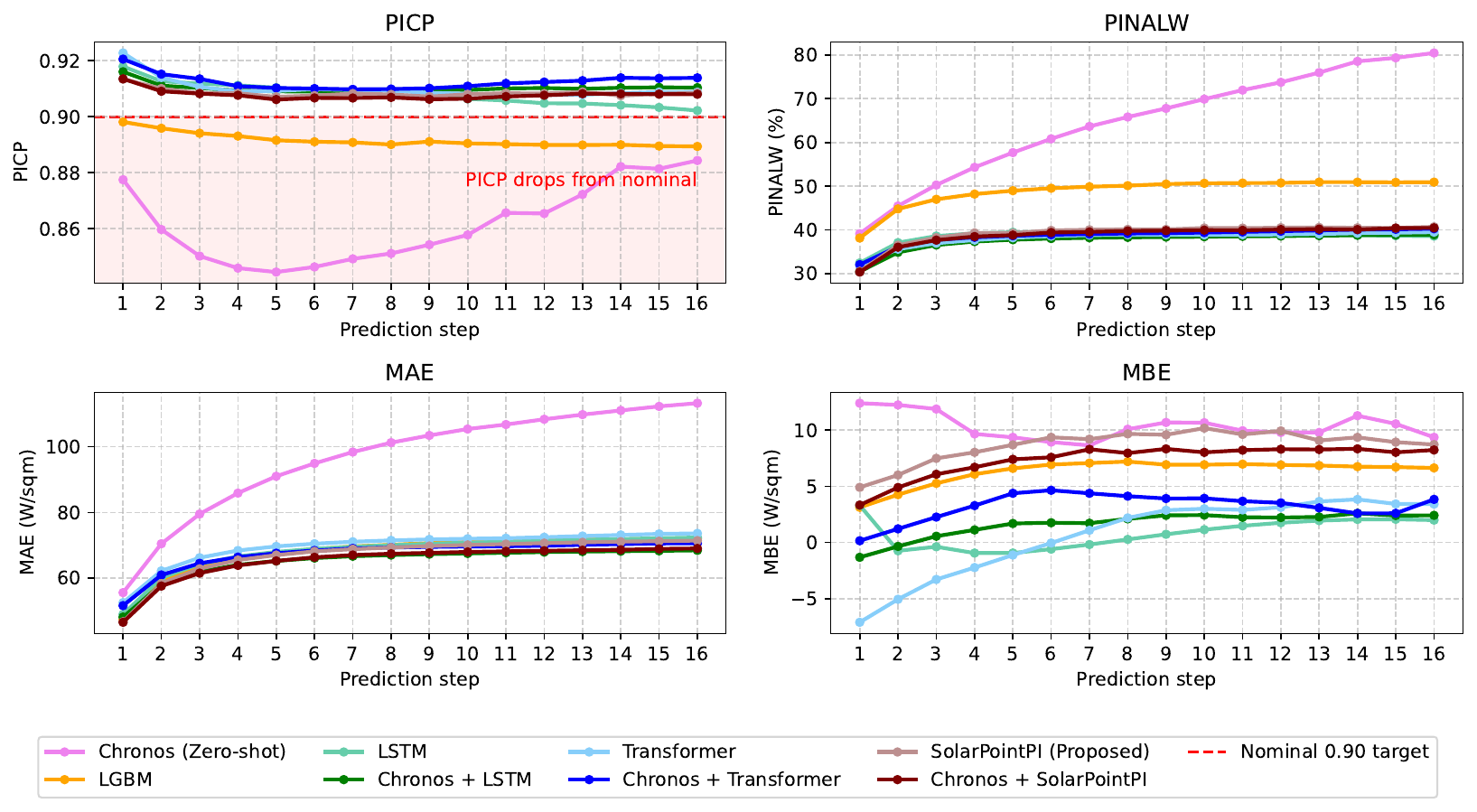}
\caption{Performance metrics comparison for model architecture benchmarking}
\label{fig:compare_model_metric}
\end{figure}

\clearpage
The performance evaluation metrics are reported in \Cref{tab:compare_model} and plotted across all 16 forecasting steps in \Cref{fig:compare_model_metric}.

\paragraph{Overall performance.} As illustrated in \Cref{fig:compare_model_metric}, the Chronos and LGBM models frequently yield PICP values below the 0.9 target coverage while producing relatively large PINALW. In contrast, remaining models trained with the proposed loss function satisfy the coverage requirement while achieving comparable PINALW and MAE. Although Chronos (a zero-shot forecaster that has not seen this dataset) provides moderately accurate point forecasts at the 15-minute horizon, its MAE increases sharply as the horizon extends, accompanied by a high MBE. While the LSTM and its Chronos-augment achieve an MBE near zero, SolarPointPI exhibits a slight residual bias.

Since all deep learning models satisfy the coverage target, we compare their PI widths in \Cref{tab:compare_model} and more closely. SolarPointPI variants achieve the lowest PINALW at the 15-minute horizon. At longer horizons, the Chronos-augmented LSTM yields the narrowest intervals, followed by the Transformer variants, which consistently rank as the second best across all lead times. PINAW results follow a similar pattern: SolarPointPI is superior for the 15-minute and 1-hour lead times, while LSTM-based models show improved results for extended horizons. Regarding the Winkler score, MAE, and RMSE, the Chronos-augmented versions of LSTM, Transformer, and SolarPointPI alternately achieve the best performance across different forecasting horizons. While the metrics remain relatively competitive between these models, Chronos+SolarPointPI tends to yield the lowest scores at shorter horizons

We find that attention mechanisms do not significantly outperform LSTM, supporting claims that Transformers struggle with strict temporal ordering \cite{Zeng2023transformer} and inter-variate impact \cite{Chen2025closerlooktransformer}. Although variants of Transformer and recurrent structure as in \cite{Kim2024transfomersolar, Chen2025transfomersolar} exist for solar power, this experiment focuses on baseline comparisons rather than architectural optimization.

\begin{table}
\centering
\caption{Performance metrics of \textbf{benchmarking models} evaluated on daytime data (06:00 - 18:00) in the test set. PICP $< 0.9$ is highlighted in \textcolor{red}{red}. Best metrics are bolded (restricted to models with PICP $\geq 0.9$). Optimal scores are the value closest to zero for MBE, and the minimum value for all other metrics. PINAW and PINALW are in \%, while MAE, RMSE, and MBE are in $\wm$. The proposed method is \textcolor{teal}{\bf SolarPointPI}.}
\small
\begin{tabular}{lccccccc}
\toprule
\textbf{Model} & \multicolumn{7}{c}{\textbf{Metrics}} \\
\cmidrule(lr){2-8}
& \textbf{PICP} & \textbf{PINAW} & \textbf{PINALW} & \textbf{Winkler} & \textbf{MAE} & \textbf{RMSE} & \textbf{MBE} \\
\toprule

\multicolumn{8}{c}{\textbf{15-minute ahead}} \\
\midrule
Chronos (zero-shot) & \color{red} 0.877 & 24.00 & 39.15 & 0.376 & 55.59 & 97.37 & 12.39 \\
LGBM & \color{red} 0.898 & 23.18 & 38.15 & 0.316 & 47.32 & 87.51 & 3.13 \\
LSTM & 0.918 & 24.75 & 32.46 & 0.393 & 49.00 & 88.14 & 3.30 \\
Chronos+LSTM & 0.916 & 22.31 & 30.47 & 0.364 & 48.14 & 86.43 & -1.32 \\
Transformer & 0.923 & 24.22 & 31.88 & 0.374 & 52.54 & 88.02 & -7.10 \\
Chronos+Transformer & 0.921 & 24.00 & 32.08 & 0.368 & 51.63 & 87.04 & \bf 0.16 \\
\textcolor{teal}{\bf SolarPointPI} & 0.913 & \textbf{20.81} & 30.55 & \bf 0.351 & 46.74 & 87.18 & 4.90 \\
Chronos+\textcolor{teal}{\bf SolarPointPI}& 0.913 & 20.99 & \textbf{30.37} & 0.352 & \textbf{46.56} & \textbf{86.32} & 3.33 \\
\midrule

\multicolumn{8}{c}{\textbf{1-hour ahead}} \\
\midrule
Chronos (zero-shot) & \color{red} 0.846 & 33.08 & 54.31 & 0.563 & 85.90 & 141.27 & 9.65 \\
LGBM & \color{red} 0.893 & 31.69 & 48.20 & 0.429 & 66.20 & 112.56 & 6.07 \\
LSTM & 0.911 & 30.58 & 39.23 & 0.497 & 66.60 & 111.33 & \bf -0.94 \\
Chronos+LSTM & 0.909 & 28.15 & \textbf{37.34} & 0.464 & 64.04 & \textbf{107.67} & 1.12 \\
Transformer & 0.909 & 29.38 & 37.69 & 0.485 & 68.42 & 111.26 & -2.23 \\
Chronos+Transformer & 0.911 & 28.96 & 38.32 & \bf 0.461 & 66.40 & 108.09 & 3.28 \\
\textcolor{teal}{\bf SolarPointPI} & 0.908 & 28.78 & 39.32 & 0.475 & 65.52 & 112.61 & 8.02 \\
Chronos+\textcolor{teal}{\bf SolarPointPI} & 0.908 & \textbf{28.00} & 38.49 & 0.462 & \textbf{63.88} & 108.99 & 6.69 \\
\midrule

\multicolumn{8}{c}{\textbf{2-hour ahead}} \\
\midrule
Chronos (zero-shot) & \color{red} 0.851 & 40.40 & 65.82 & 0.646 & 101.25 & 161.52 & 10.07 \\
LGBM & \color{red} 0.890 & 33.88 & 50.13 & 0.466 & 70.15 & 117.20 & 7.19 \\
LSTM & 0.908 & 31.22 & 39.45 & 0.523 & 70.04 & 115.27 & \bf 0.27 \\
Chronos+LSTM & 0.910 & \textbf{29.22} & \textbf{38.30} & 0.485 & \textbf{67.02} & \textbf{111.37} & 2.10 \\
Transformer & 0.909 & 30.70 & 38.91 & 0.507 & 71.49 & 114.83 & 2.19 \\
Chronos+Transformer & 0.910 & 29.87 & 39.19 & \bf 0.481 & 69.32 & 111.64 & 4.12 \\
\textcolor{teal}{\bf SolarPointPI} & 0.908 & 30.37 & 40.14 & 0.509 & 69.32 & 116.98 & 9.66 \\
Chronos+\textcolor{teal}{\bf SolarPointPI} & 0.907 & 29.40 & 39.77 & 0.485 & 67.43 & 113.11 & 7.95 \\
        \midrule

\multicolumn{8}{c}{\textbf{4-hour ahead}} \\
\midrule
Chronos (zero-shot) & 0.884 & 50.36 & 80.46 & 0.712 & 113.26 & 176.93 & 9.35 \\
LGBM & \color{red} 0.889 & 34.97 & 50.93 & 0.479 & 71.97 & 118.78 & 6.63 \\
LSTM & 0.902 & 30.97 & \textbf{38.58} & 0.537 & 72.37 & 117.01 & \bf 1.99 \\
Chronos+LSTM & 0.910 & \textbf{29.76} & 38.89 & 0.493 & \textbf{68.45} & 112.82 & 2.41 \\
Transformer & 0.909 & 31.49 & 39.49 & 0.518 & 73.62 & 116.67 & 3.41 \\
Chronos+Transformer & 0.914 & 30.96 & 40.28 & \bf 0.485 & 70.60 & \textbf{112.24} & 3.83 \\
\textcolor{teal}{\bf SolarPointPI} & 0.908 & 31.41 & 40.71 & 0.525 & 71.42 & 118.66 & 8.71 \\
Chronos+\textcolor{teal}{\bf SolarPointPI} & 0.908 & 30.36 & 40.46 & 0.499 & 69.03 & 114.99 & 8.22 \\
\bottomrule
\end{tabular}
\label{tab:compare_model}
\end{table}

\clearpage
\begin{figure}
\centering
\includegraphics[width=0.7\textwidth]{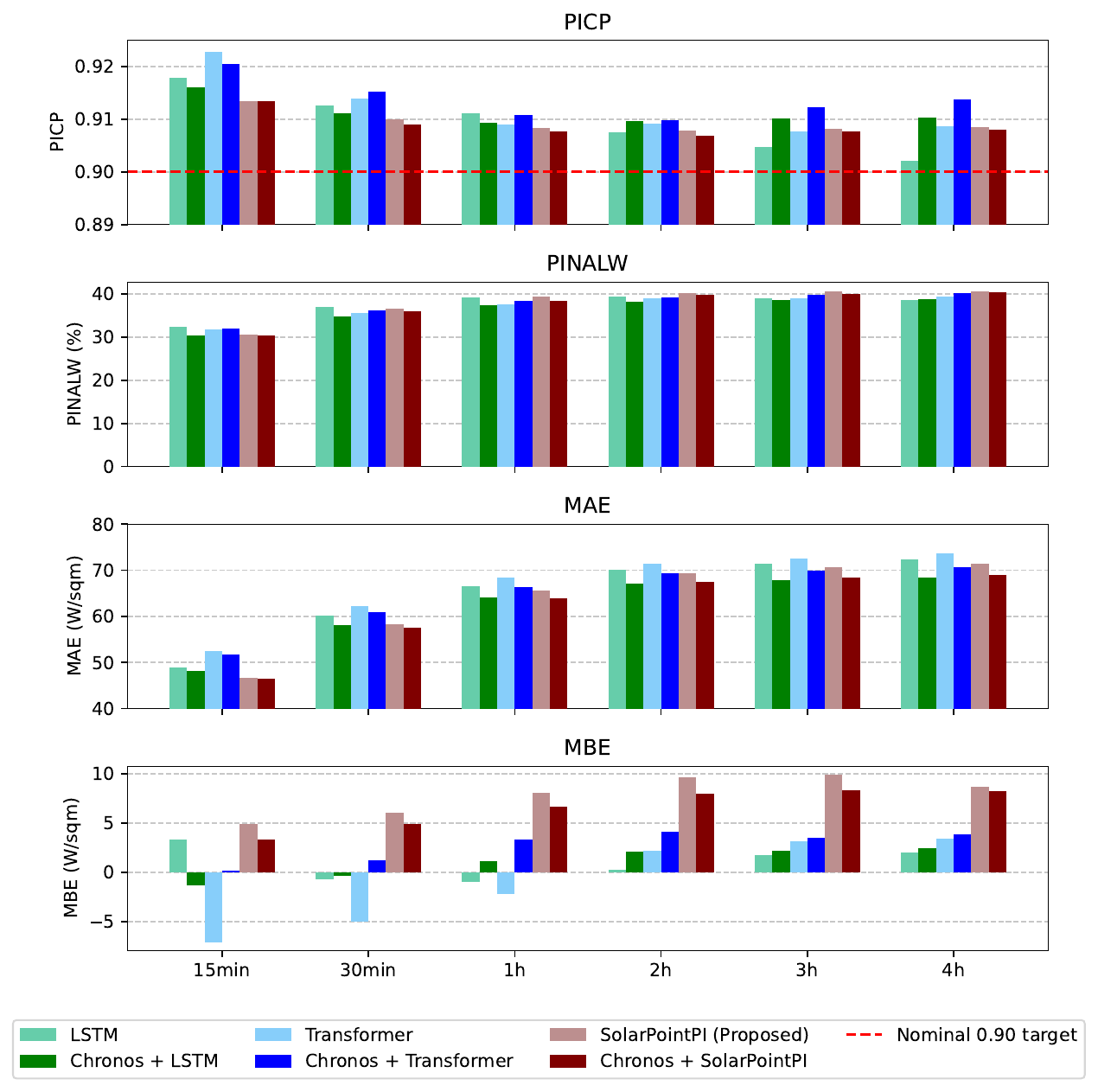}
\caption{Performance comparison of base models with and without Chronos 2.0}
\label{fig:comparison-base-chronos}
\end{figure}

\paragraph{Effect of Chronos 2.0 feature augmentation.} \Cref{fig:comparison-base-chronos} illustrates the impact of utilizing Chronos 2.0 as a feature generator across the three base architectures. All model variants maintain a valid PICP above the 0.90 target level. Notably, augmenting SolarPointPI with Chronos features reduces the PINALW, MAE, and MBE across all prediction steps, suggesting that Chronos features facilitate narrower prediction intervals and more accurate point forecasts without compromising reliability. For the LSTM model, the inclusion of Chronos leads to a significant reduction in all metrics across nearly all horizons, with the exception of the 4-hour horizon, where the PINALW and MBE experience slight degradation. Conversely, the benefit of Chronos for the Transformer architecture is primarily reflected in the MAE and MBE; unexpectedly, the PINALW for the Chronos+Transformer configuration increases relative to the base model.

Ultimately, whether the improvement gained by adding Chronos is statistically or practically significant depends heavily on the specific base architecture, the choice of metrics, and the forecast horizons. Consequently, selecting a superior model may require considering additional factors, such as computational complexity and numerical implementation requirements.

\begin{figure}
\centering
\includegraphics[width=\textwidth]{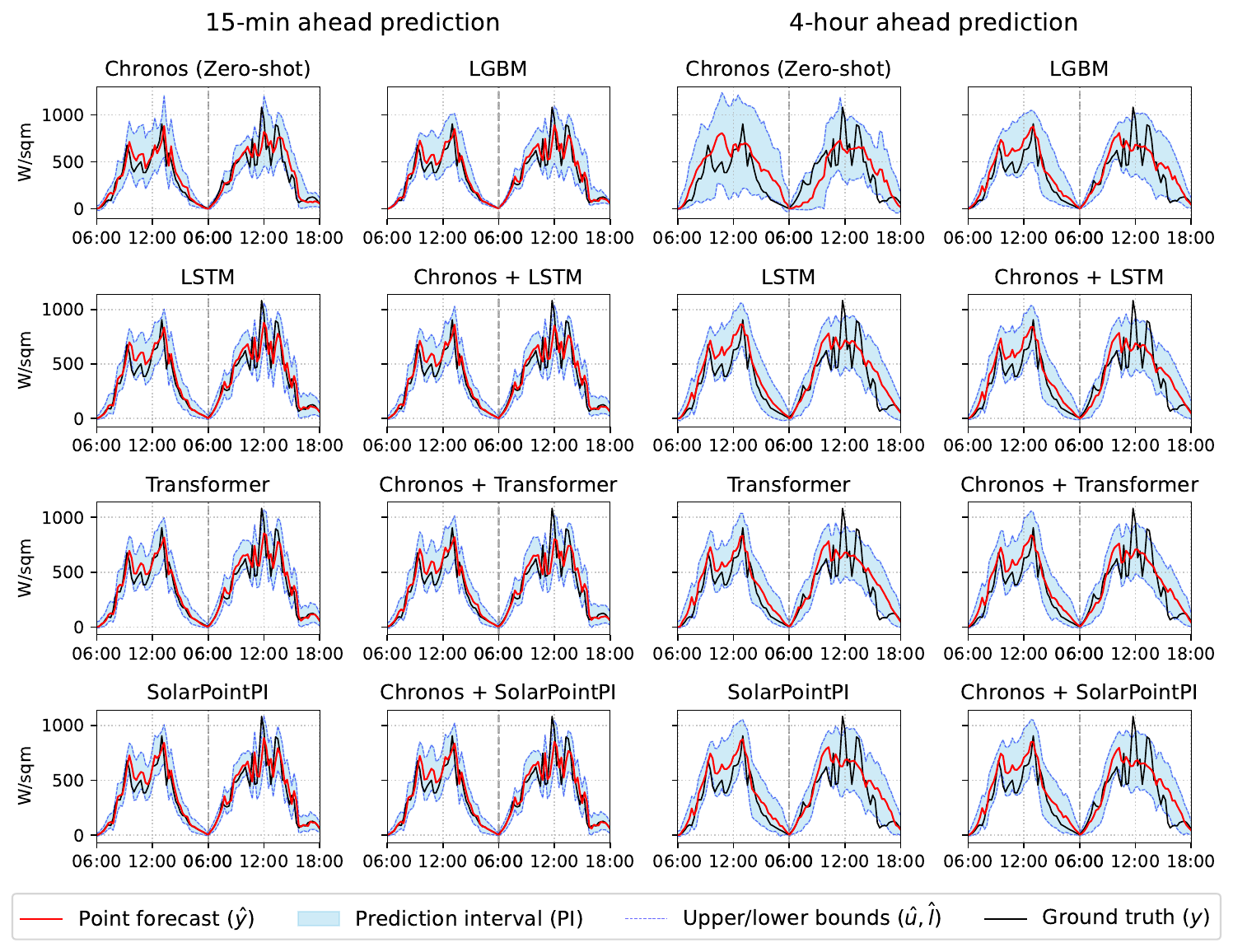}
\caption{Solar irradiance forecasts (W/sqm) comparing model architectures at 15-minute and 4-hour prediction horizons under cloudy-sky conditions.}
\label{fig:compare_model_ts}
\end{figure}

\clearpage
\paragraph{Example of forecasting result.}
\Cref{fig:compare_model_ts} illustrates the performance of various models at 15-minute and 4-hour horizons under cloudy sky conditions. The zero-shot Chronos model exhibits a very high uncertainty estimate, resulting in extremely wide PIs at both horizons. Similarly, the LGBM model produces relatively large intervals to capture sudden irradiance spikes. In contrast, the LSTM, Transformer, and SolarPointPI models show a tighter fit to the ground truth, with visually comparable PI widths and point forecast accuracy. All methods produce wider PIs at the 4-hour horizon, as uncertainty naturally increases as the prediction horizon extends.

\subsection{Computational complexity}
\begin{figure}
    \begin{minipage}[c]{0.48\textwidth}
        \centering
        \includegraphics[width=\linewidth]{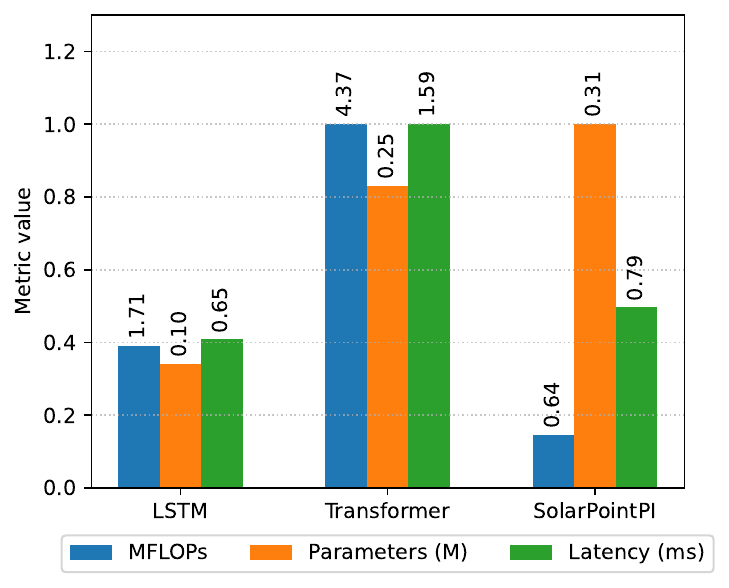}
        \caption{Computational complexity.}
        \label{fig:comp_complexity}
    \end{minipage}\hfill
    \begin{minipage}[c]{0.48\textwidth}
    \small
        \centering
        \captionof{table}{Computational complexity comparison.}
        \label{tab:computational_complexity}
        \begin{tabular}{lccc}
        \toprule
         \textbf{Model} & \textbf{MFLOPs} & \textbf{Parameters (M)} & \textbf{Latency (ms)} \\
        \midrule
        Chronos & 3246.600 & 119.478 & 7.323 \\
        (zero-shot) &&& \\
        LSTM & 1.708 & 0.104 & 0.649 \\
        Transformer & 4.370 & 0.254 & 1.590 \\
        SolarPointPI & 0.642 & 0.306 & 0.789 \\
        \bottomrule
        \end{tabular}
    \end{minipage}
\end{figure}

All benchmarks were conducted on an Ubuntu 24.04 LTS system equipped with an Intel Core i7-14700 processor (20 cores, 28 threads), an NVIDIA GeForce RTX 5070 Ti GPU with 16 GB of VRAM, and 64 GB of RAM, using Python 3.13.11 within PyTorch implementation \cite{Pytorch2019}. To evaluate model complexity, the \texttt{ptflops} library \cite{ptflops} was used to calculate the Multiply-Accumulate operations (MACs) and total parameter count, with the resulting MACs multiplied by approximately two to estimate total FLOPs. Inference latency was benchmarked over 100 iterations following a 10-iteration warmup phase to stabilize hardware caches. All inference measurements were performed on the test set with a batch size of 1, reflecting the practical deployment scenario for solar forecasting, where the model predicts a single sample at a time.

As shown in \Cref{fig:comp_complexity} and \Cref{tab:computational_complexity}, the computational complexity analysis demonstrates the efficiency of the proposed SolarPointPI model benchmarking with other model architectures.
 
SolarPointPI achieves the lowest FLOP count at 0.642 MFLOPs. The LSTM records a higher count of 1.708 MFLOPs, due to its encoder-decoder structure that requires two sequential passes through the network. SolarPointPI avoids this by using a single-pass architecture, and its lightweight output heads further contribute to the reduction in arithmetic operations. The Transformer has the highest FLOP count among the compact models at 4.370 MFLOPs, a result of its $\mathcal{O}(n^2)$ attention complexity where $n$ is the sequence length. This
quadratic scaling becomes particularly costly at batch size 1, where the overhead of the attention mechanism is not distributed across samples. SolarPointPI has more parameter counts but fewer FLOPs than LSTM because it pairs a single LSTM with separate, submodules for each output step. In contrast, the LSTM (encoder-decoder) reduces parameters by sharing a single output head, but its dual-LSTM encoder-decoder architecture significantly increases overall computational complexity.

In terms of inference latency, SolarPointPI completes a forecast in 0.789 ms, which is comparable to the LSTM at 0.649 ms and substantially lower than the Transformer at 1.590 ms. These results confirm that the efficiency gains from the single-pass design translate directly to reduced runtime. The inference time increases significantly when considering the Chronos (zero-shot) baseline, which requires 3,246.6 MFLOPs and 119.478 M parameters. In Chronos-augmented variants, inference time is dominated by this pretrained backbone; even with frozen weights, the full parameter set must be evaluated at each step. Nevertheless, total inference time remains in the millisecond range, making real-time deployment feasible. Given the resulting performance gains, the additional latency introduced by Chronos is a justifiable trade-off.

\section{Conclusions}
\label{sec:conclusion}
This paper demonstrates the effectiveness of a unified NN framework for simultaneous point and interval forecasting, driven by a novel PI loss and a multi-objective training strategy using multiple gradient descent directions. We successfully addressed the fundamental trade-off between target coverage (reliability) and interval width (sharpness) by employing an extended log-barrier function for the former and the Sum-$k$ loss for the latter. Experimental results in intra-day solar irradiance forecasting--a domain characterized by high fluctuation--confirm that our method consistently achieves the target PICP with superior sharpness compared to existing scalarized loss functions. Furthermore, our approach eliminates the intensive hyperparameter tuning typically required in literature. The framework's versatility was validated by integrating it with state-of-the-art architectures, including LSTM encoder-decoders, Transformers, and Chronos-augmented models, showing competitive performance across all configurations. While this work focuses on prediction intervals, the framework is architecturally flexible and can be adapted to different input features or more complex models depending on data availability for real-time deployment. Future research will explore extending this multi-objective optimization approach to other forms of uncertainty quantification, such as cumulative distribution functions (CDFs) or multi-quantile estimation, to support a broader range of risk-based decision-making tasks.

\section{Acknowledgment}
\label{sec:acknowledgement}

This research project is financially supported by the Ratchadaphiseksomphot Endowment Fund, Chulalongkorn University. During the preparation of this work the authors used Gemini 3 Flash in order to improve readability and language. After using this tool, the authors reviewed and edited the content as needed and took full responsibility for the content of the published article.

\clearpage
\bibliographystyle{alpha}
\bibliography{pipointref}
\addcontentsline{toc}{section}{References}


\end{document}